# Towards the automated large-scale reconstruction of past road networks from historical maps

Johannes H. Uhl [1,2,*], Stefan Leyk [2,3], Yao-Yi Chiang [4], and Craig A. Knoblock [5,6]

[1]Earth Lab, Cooperative Institute for Research in Environmental Sciences, University of Colorado Boulder, Boulder, CO 80309, USA
[2]Institute of Behavioral Science, University of Colorado Boulder, Boulder, CO 80309, USA
[3]Department of Geography, University of Colorado Boulder, Boulder, CO 80309, USA
[4]Department of Computer Science & Engineering, University of Minnesota, Minneapolis, MN 55455, USA
[5]Information Sciences Institute, University of Southern California, Marina del Rey, CA 90292, USA
[6]Spatial Sciences Institute, University of Southern California, Los Angeles, CA 90089, USA

*Corresponding author (johannes.uhl@colorado.edu), Cooperative Institute for Research in Environmental Sciences, 216 UCB, Boulder, CO 80309, USA.

**Abstract:** Transportation infrastructure, such as road or railroad networks, represent a fundamental component of our civilization. For sustainable planning and informed decision making, a thorough understanding of the long-term evolution of transportation infrastructure such as road networks is crucial. However, spatially explicit, multi-temporal road network data covering large spatial extents are scarce and rarely available prior to the 2000s. Herein, we propose a framework that employs increasingly available scanned and georeferenced historical map series to reconstruct past road networks, by integrating abundant, contemporary road network data and color information extracted from historical maps. Specifically, our method uses contemporary road segments as analytical units and extracts historical roads by inferring their existence in historical map series based on image processing and clustering techniques. We tested our method on over 300,000 road segments representing more than 50,000 km of the road network in the United States, extending across three study areas that cover 53 historical topographic map sheets dated between 1890 and 1950. We evaluated our approach by comparison to other historical datasets and against manually created reference data, achieving F-1 scores of up to 0.95, and showed that the extracted road network statistics are highly plausible over time, i.e., following general growth patterns. We demonstrated that contemporary geospatial data integrated with information extracted from historical map series open up new avenues for the quantitative analysis of long-term urbanization processes and landscape changes far beyond the era of operational remote sensing and digital cartography.

**Keywords:** Historical maps, topographic map processing, spatial data integration, road network analysis, transportation infrastructure, land development, urbanization, historical GIS.

## 1. Introduction

Road networks are an important component of international, regional, and local transportation infrastructure and are the backbone of economy, trade, and tourism. They reflect settlement and land development patterns, alongside historical migration and trading routes, or railroad networks and thus, seen from a historical network perspective, represent a physical documentation of the dynamics of our civilization (Jacobson 1940). In an urban context, road networks reflect different phases of urban growth (Boeing 2020), and determine potential walkability (Gori et al. 2014), efficiency (Merchan et al. 2020), and sustainability (Rao et al. 2018) of cities. Thus, they represent a driver of socio-economic urban processes (Graham & Marvin 2002) related to accessibility (Coppola & Papa 2013), commutability, public health (Frizzelle et al. 2009), and equity (Santos et al. 2008), not only related to automobile-based transportation but also in the context of public transport (Daniels & Mulley 2013).

Detailed information on contemporary road networks and their geometric, semantic, material-related and dynamically changing properties such as utilization load, traffic conditions etc., can be acquired from remote sensing data (e.g., Schnebele et al. 2015, Zhang et al. 2019, Palubinskas et al. 2009), through crowd-sourced data (e.g., Barrington-Leigh & Millard-Ball 2017, Boeing 2017) or user-collected data harvested from navigation apps or devices (Cohn 2009, Tang et al. 2012). Moreover, geometric changes in recent decades can be detected and quantified using historical GIS data (Zhang & Couloigner 2005). However, surprisingly little spatially explicit information is available on the evolution of urban and rural road networks over extended periods of time, which is crucial to fully understand the evolution of transportation infrastructure, enabling more informed urban and regional planning (Levinson 2005).

However, data even on the geometric properties of road networks prior to the 1980s are scarce. The few existing quantitative, long-term studies revolving around transport infrastructure evolution typically rely on manually vectorized road network data (Strano et al. 2012, Masucci et al. 2013, Masucci et al. 2014, Casali & Heinimann 2019, Wang et al. 2019, Kaim et al. 2020, El Goui & Lagesse 2021) or railroad network data (Thévenin et al. 2013, Donaldson & Hornbeck 2016), from multi-modal data sources, often involving labor-intensive manual digitization work.

Consequently, long-term studies on the evolution of road networks over large spatial extents and at fine spatial grain are scarce. The few existing approaches use ancillary data such as multi-temporal urban-rural classifications at the census tract level (Boeing 2020), building age information at the cadastral parcel level (Barrington-Leigh & Millard-Ball 2015), as well as multi-temporal gridded settlement surfaces derived from parcel-level records (Millard-Ball 2021, Burghardt et al. 2021) or from remote sensing observations (Barrington-Leigh & Millard-Ball 2020). These approaches infer the age of roads based on the age of nearby buildings or based on the development period of the surrounding areas and are typically constrained to local roads within developed land, omitting roads in undeveloped, rural areas.

To overcome this lack of historical road network data, researchers dedicated to the field of topographic map processing (Chiang et al. 2014, Liu et al. 2019) have developed methods for the extraction of road network features (or their components, such as road intersections) from historical maps, and from topographic maps in general, using image processing, template matching, classification, and image segmentation techniques.

For example, Bin and Cheong (1998) use connected component analysis and a vectorization technique, whereas Callier and Saito (2011) use linear feature detection in combination with a region growing algorithm to extract road geometries from scanned maps. Itonaga et al. 2003 use a stochastic relaxation algorithm and a thinning operator, while Chiang et al. (2005) employ histogram-based segmentation and parallel pattern tracing for the same purpose. Similar approaches use color clustering (Chiang & Knoblock 2009) or morphological operations (Chiang & Knoblock 2008) to extract road features and road intersections. Despite being unsupervised approaches, most of these methods require some user interaction, e.g., to determine which cluster represents road features. Other examples are supervised and are based on localized template matching (Chiang et al. 2009) or on Hough transform in combination with an edge matching algorithm (Chiang & Knoblock 2013).

More recently, scholars have applied deep learning methods such as convolutional neural networks (CNNs) to extract geometric and semantic transportation network characteristics from historical maps. These approaches include linear road feature extraction based on a U-Net CNN (Ekim et al. 2021), extraction of road network intersections using an Inception-ResNet CNN (Saeedimoghaddam & Stepinski 2020), or road type recognition from cartographic road symbols using a U-Net CNN (Can et al. 2021). Similarly, researchers have proposed deep learning based methods for the extraction of railroad networks (Hosseini et al. 2021, Chiang et al. 2020a) from historical maps. These deep-learning based methods are resource-intensive and require large amounts of typically manually labelled training data or templates. Jiao et al. (2021) provide a detailed overview of these methods.

We propose an alternative, fully automated approach, making use of abundantly available, contemporary geometric road network data, in combination with image processing and unsupervised classification techniques applied to digital historical maps. This approach is based on the assumption that road networks typically expand over time (rather than shrink or experience other types of changes), and thus, the contemporary road network represents the superset of all roads being depicted in the historical maps. The proposed method aims to separate contemporary road network vector data in two classes: those roads that exist in an underlying historical map (i.e., historical roads) and those that do not exist in that map (i.e., more recent roads). This separation is done in an unsupervised manner, and thus, no labelled training data is required. Moreover, in contrast to most existing map processing approaches, we do not use the pixels of the scanned map image as analytical units, but rather the contemporary (vector) road segments, typically representing the center line of the roads. Hence, our approach filters the already topologically cleaned, contemporary road vector data based on signals extracted from the color information harvested from historical maps, and thus, avoids the complex, and potentially error-prone recognition (e.g., Chiang et al. 2020a) and vectorization (e.g., Chen et al. 2021) of cartographic content in historical maps (i.e., road symbols). It thus enables spatial, spatio-temporal, and network-based retrospective analyses using the contemporary road segments as analytical units.

The existing supervised and unsupervised road extraction methods typically require a considerable degree of user interaction (e.g., manual labelling of training data, or parameter tweaking), and most of the methods have only been tested on individual map sheets. Thus, it remains unclear how these methods perform on large, potentially heterogeneous map collections. Moreover, existing approaches do not incorporate contemporary road network data to guide the extraction. Thus, our proposed method makes the following contributions: (a) it is a fully automated approach to extract road networks from historical maps, (b) it is the first approach using vector-raster data integration (i.e., combining contemporary road network data and scanned historical maps), and (c) it has been tested over several, large study areas, and for different time periods. Furthermore, the proposed approach requires very few parameters to be set by the user, and the results are largely invariant to the choice of these parameters, as we will show herein.

This effort is motivated by the growing availability of systematically scanned, georeferenced, and catalogued historical map archives, increasingly available as public and open data (Fishburn et al. 2017, Library of Congress 2020, National Library of Scotland 2020, Swisstopo 2020, Stanford University Library 2020, Biszak et al. 2017, Old Maps Online 2020, see Uhl & Duan 2021, McDonough 2022). At the time of writing, the number of scanned and / or georeferenced historical maps from national map archives available online is expected to exceed 1,000,000 (McDonough 2022), and unlocking the unique, historical-spatial information contained in these map archives (i.e., extracting map content and converting it into analysis-ready spatial data structures) constitutes the overall goal of topographic map processing (Chiang et al. 2014). Moreover, there is an increasing demand of historical spatial data for numerous applications in urban studies and planning (Dunne et al. 2016), as well as in the digital humanities (Gregory and Healey 2007, Chiang et al. 2020b, Hosseini et al. 2021).

Herein, we apply our method to a range of historical topographic maps from the United States (Section 2.1) and present the details of this method (Section 2.2). We implemented several strategies for validation, cross-comparison, and plausibility testing to evaluate our results (Section 2.3). We show the results of our analyses in Section 3, we discuss them in Section 4, and conclude with a critical reflection and an outlook on future work (Section 5).

## 2. Data & Methods

Herein, we describe a method that estimates for each contemporary road network segment whether a corresponding road symbol exists in a given scanned and georeferenced historical map of the same area. This is done by (a) using an image processing-based, continuous metric that indicates the likelihood that such a road symbol exists, and (b) employing a discretization method to convert this continuous metric into a binary metric indicating the presence or absence of a road symbol at the corresponding location on a given historical map.

In this section, we describe the acquisition of historical maps for three study areas, located in the US, and six different points in time, ranging from 1895 to 1950 (Section 2.1). We then detail the characteristics of the contemporary road network vector data, which we acquired and processed for these study areas (Section 2.2) and present a manual and an automatic strategy to generate validation data in order to evaluate our approach (Section 2.3). We then describe the image processing pipeline to generate continuous estimates of historical road existence (Section 2.4), and the subsequent discretization step to extract historical road segments from the pool of the entire contemporary road network (Section 2.5). Finally, we present four different strategies that we employed to test the performance of the proposed method and the plausibility of the results (Section 2.6).

### 2.1. Historical map acquisition

We obtained historical maps from the United States Geological Survey (USGS) historical topographic map collection (HTMC), which is a digital archive of more than 190,000 scanned and georeferenced topographic maps created between 1884 and 2006 (Allord et al. 2014). Specifically, we used metadata for the HTMC (available from https://thor-f5.er.usgs.gov/ngtoc/metadata/misc/) to generate the geographic footprints of each map sheet contained in the archive, and obtained historical map sheets for a range of U.S. metropolitan areas, by automatically downloading them from the Amazon Web Services (AWS) S3 archive where the HTMC is hosted (USGS 2021). We then inspected the temporal and geographic coverage of the maps in each metropolitan area, examining the two largest map scales (i.e., 1:24,000 and 1:62,500) and chose three metropolitan areas with complete coverage for one or more (up to three) early time periods. The maps from different areas often exhibit different cartographic styles representing various geographic settings. We discarded the 1:24,000 scale maps, as they tend to be more recent than the 1:62,500 maps (see Uhl et al. 2018). The study areas are Greater Albany (New York), for the approximate years 1900, 1930, and 1950, consisting of six map quadrangles (Fig. 1 a-c), 11 map quadrangles for the San Francisco Bay area (California) in approximately 1900 and 1950 (Fig. 1 d,e), and a study area covering four map quadrangles in the Mobile Bay (Alabama) in approximately 1920. Henceforth, we call these combinations of study areas and time periods NY-1900, NY-1930, NY-1950, CA-1900, CA-1950, and AL-1920, respectively. In total, we used 53 different map sheets, covering a range of color tones and contrast levels, as well as different printing techniques (e.g., black and white print in the AL-1920 study area, 5-color print in the CA-1950 study area). For each study area, we used an automated procedure to (a) remove the map collars, and (b) generate seamless mosaicked layers. This procedure has been developed for a previous project focusing on the extraction of urban areas and is described in detail in Uhl et al. (2021a).

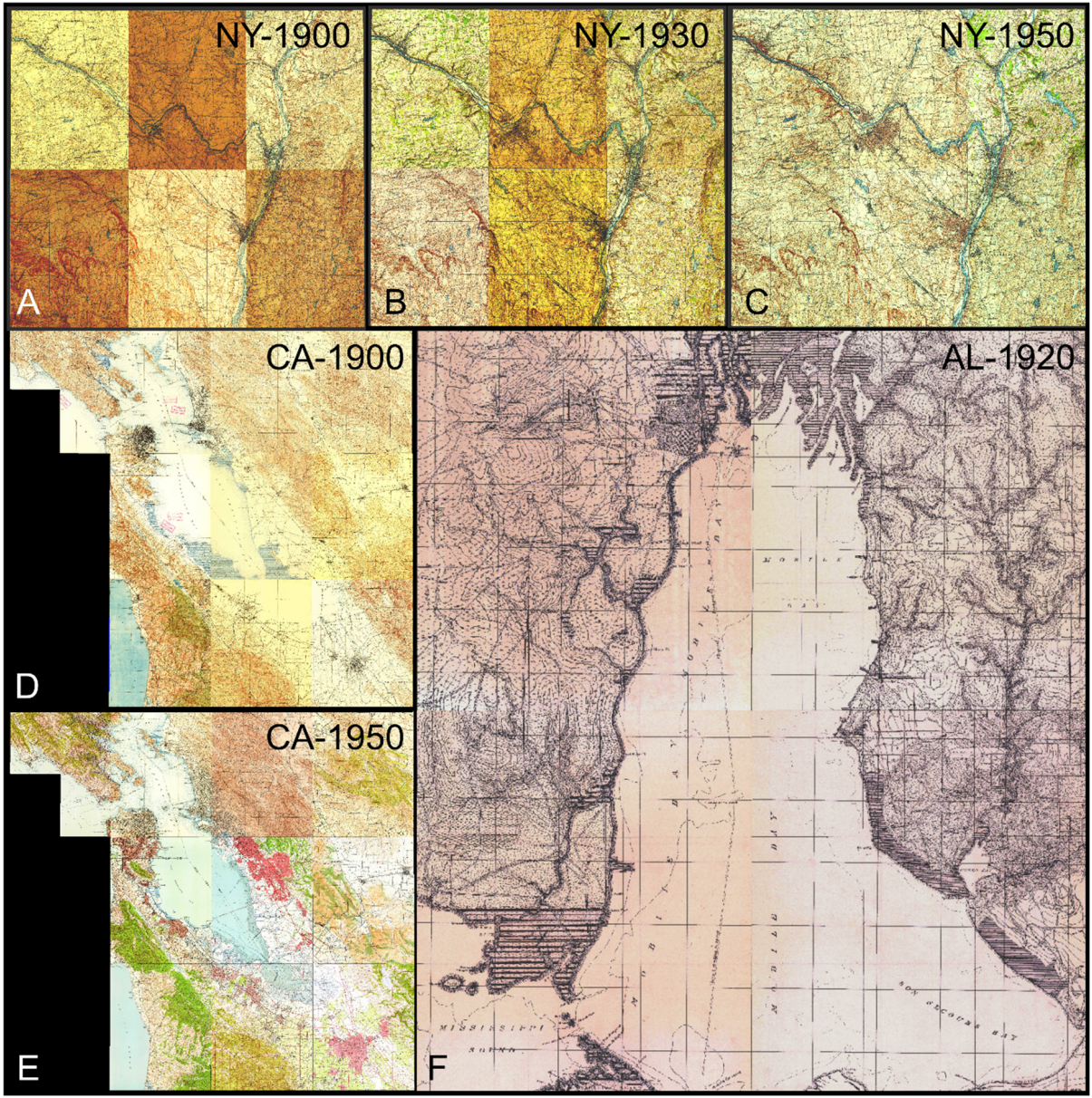

**Figure 1. Historical 1:62,500 topographic maps of the three study areas. A composite of six map sheets covering Greater Albany (New York) in (a) 1893, (b) 1930, and (c) 1950; a 11-map composite covering the Bay Area, California in (d) 1895 and (e) 1950, and (f) a four-map composite covering Mobile Bay, Alabama, in 1920.**

In contrast to these differences in general map appearance, the way how roads are depicted appears to be fairly homogeneous across time periods and cartographic styles. As can be seen in Fig. 2, all study areas use parallel black lines to depict streets, in some cases generalized to the street blocks, or merged with building blocks or individual building outlines. In the 1950 maps, dense urban areas are depicted using red dots (NY-1950, Fig. 2c) or in a pink color signature (CA-1950, Fig. 2e) underlying the road symbols.

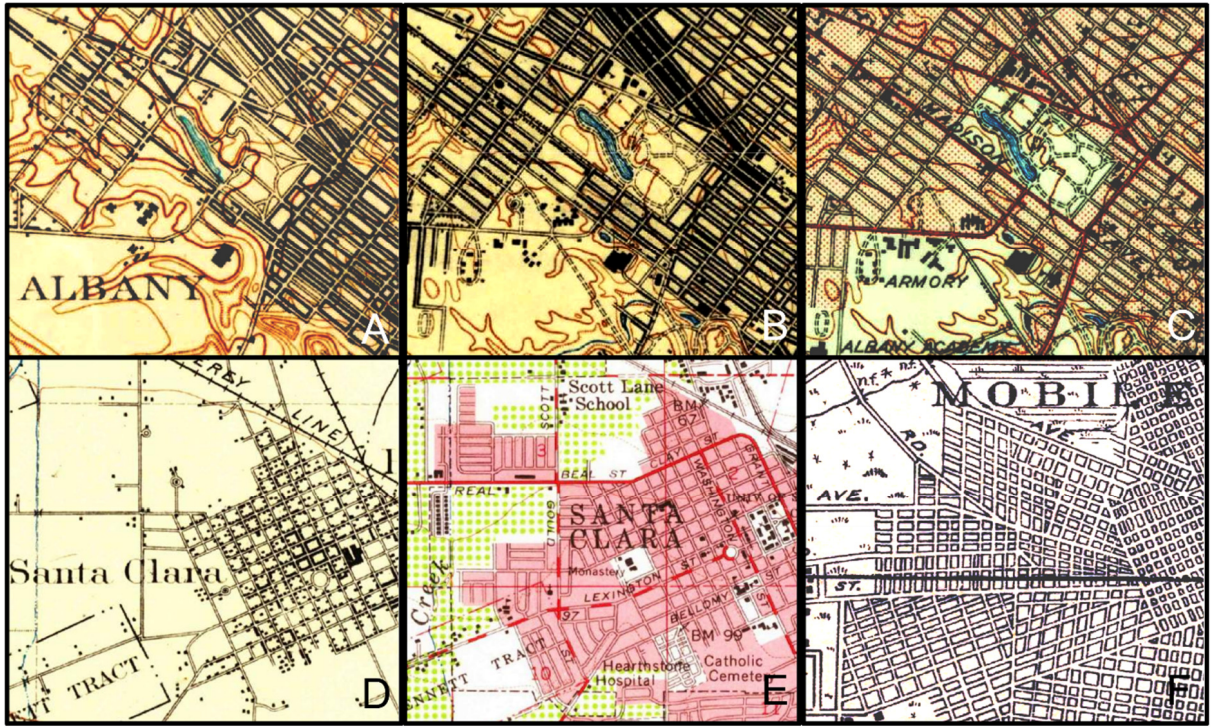

**Figure 2.** Cartographic styles used for road depiction in the historical maps. Albany (New York) in (a) 1900, (b) 1930, and (c) 1950; Santa Clara (California) in (d) 1895 and (e) 1950, and (f) Mobile (Alabama) in 1920.

## 2.2. Road network data preprocessing

In addition to the historical maps from the HTMC, we e used the National Transportation Dataset (NTD, v2019, USGS 2019) from the USGS as contemporary road network data. The NTD is available as geospatial vector data, containing several feature classes on road and railroad networks, per state. The road network feature classes are generally topologically clean, i.e., an individual line feature represents a straight or a curved road (i.e., the road centerline) between two intersections, or between a dead-end and an intersection. Herein, we refer to these linear features as "road segments". An actual street, as defined by a street name, may consist of multiple road segments. We clipped the road network vector data to the extents of the three study areas shown in Fig. 1. As we expect our results to vary across rural-urban gradients, we stratify the NTD road segments into two classes, assuming that short road segments are likely to occur in dense, urban areas, and long road segments typically occur in sparsely settled rural areas. Thus, we stratify the road segments into short ("urban") roads and long ("rural") roads, based on the 90$^{th}$ percentile of the road segment length distributions per study area as the threshold. This threshold corresponds to absolute values ranging between 345m and 469m across study areas (Table 1). The effect of this stratification is shown in Fig. 3a,b; see Appendix Fig. A1 for a visual evaluation of this stratification against building density estimates. Table 1 shows some basic statistics on the road networks in each of the three study areas.

**Table 1. Basic road network statistics in the three study areas.**

| Study area | Time periods per study area | Total road segments | Urban road segments | Rural road segments | Total km road | km urban roads | km rural roads | Urban-rural threshold (90th percentile) [m] |
|---|---|---|---|---|---|---|---|---|
| AL | 1920 | 41,494 | 37,344 | 4,150 | 6,909 | 4,320 | 2,589 | 345 |
| CA | 1900, 1950 | 222,517 | 200,265 | 22,252 | 35,896 | 22,430 | 13,465 | 318 |
| NY | 1900, 1930, 1950 | 45,354 | 40,818 | 4,536 | 10,078 | 5,851 | 4,228 | 469 |

## 2.3. Validation data generation

To our knowledge, there is no vector-based, multi-temporal road network data covering the study periods used herein that would be (a) compiled independently from the data under test, (b) presumably of the same or higher levels of accuracy, and (c) represent a large enough sample to generate accuracy estimates of high statistical power (Congalton & Green 2019), and thus could be used as reference data. To overcome this issue, we used a two-fold strategy to generate reference data as follows.

### 2.3.1. Manually labeled patch-level validation data

First, we took a stratified random sample (N=100) of rural and urban roads per study area and year, summing up to a total of 1,200 road segments. For each road segment, we cropped the historical maps within a patch of 500 m × 500 m around the segment centroid, and manually annotated these patches; we assigned a binary label indicating the presence or absence of a road symbol in the approximate center of the map patch. The random sampling yielded a relatively balanced dataset, i.e., 58% positive (road present), 42% negative (no road present) labels.

### 2.3.2. Automatically created building-based validation data

As the sample of N=1,200 only covers 0.2% of the overall set of 627,909 observations (i.e., road segments per study area and time period, see Table 1), we also implemented a procedure that annotates each road segment with a reference label. This procedure is based on the assumption that if there is a road at a given location in a historical map it is likely that one or more buildings would have existed somewhere in proximity along the road segment. While the co-evolution of roads and buildings is little studied (Achibet et al. 2014), we assume that this expectation is reasonable for most roads in urban settings, and for a fair amount of roads in rural settings.

We use historical built-up area (BUA, Uhl & Leyk 2020) surfaces from the Historical Settlement Data Compilation for the US (HISDAC-US, Leyk & Uhl 2018, Uhl et al. 2021b), which are available in 5-year intervals for the time period from 1810 to 2016 and are derived from built year information contained in the Zillow Transaction and Assessment Dataset (ZTRAX, Zillow 2021). The binary BUA surfaces measure the presence of at least one built-up property within grid cells of 250m × 250m, in a given year, and have been employed for long-term studies of the built environment (Leyk et al. 2020, Uhl et al. 2021c) and, under similar assumptions, for historical road network modeling (Boeing 2020, Millard-Ball 2021). Fig. 3c-h shows the historical BUA layer sequences from the HISDAC-US, for each study area, and for the half-decade closest to each study period.

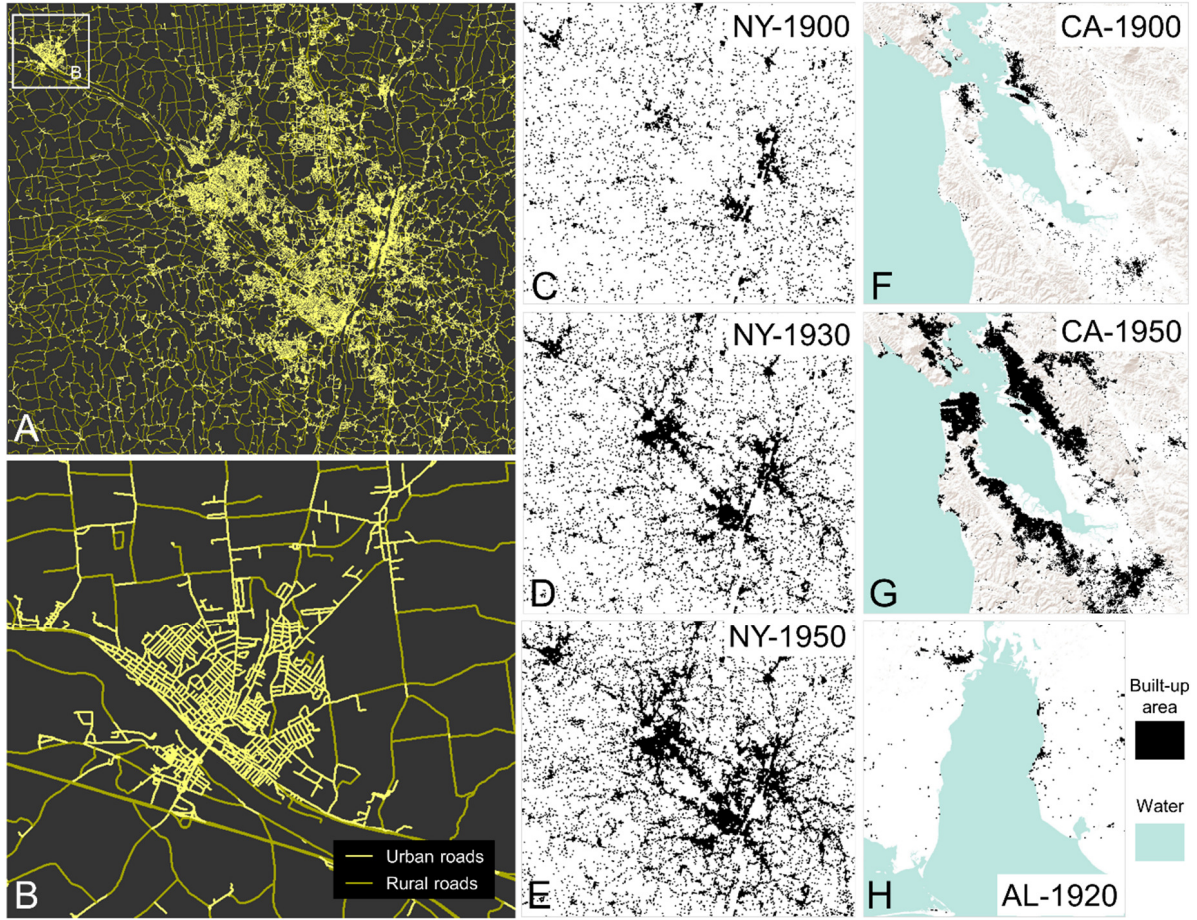

**Figure 3. Road network and validation data.** (a) Contemporary NTD road network data, stratified into urban (short) and rural (long) road segments, shown for the whole New York study area, and (b) for the city of Amsterdam (New York); (c)-(e): Historical built-up areas (BUA) from the HISDAC-US dataset, derived from building construction year information from ZTRAX, originally obtained from county-level tax and assessment data, shown for the New York study area in 1900, 1930, and 1950, respectively, (f),(g) BUA for the California study area in 1900, and 1950, and (h) for the Alabama study area in 1920. Grid cells labelled as "built-up area" contain at least one built-up property in a given year. Hillshade source in (c)-(h): World Terrain Base (Esri, USGS, NOAA).

Based on these historical built-up area delineations, we estimate the historical built-up area in proximity to each road segment of the contemporary NTD road network data (Fig. 4a), as follows: We buffer each road segment by 125m (= half of the HISDAC-US grid resolution) (Fig. 4b). For the buffered NTD road segments within each map sheet extent (from a map referenced to year $T$), we extract the HISDAC-US BUA layer (Fig. 4c) for the most recent half-decade that is <$T$ (Fig. 4d). That is, for a map sheet from 1899 we use the HISDAC-US BUA layer from 1895, in order to include buildings that have been established by the time of the survey underlying the historical map. Next, we calculate the fraction of built-up grid cells within each buffer area. This fraction quantifies how much of the area in proximity to each road segment has been built-up in the year $T$. Finally, we append this fraction to the original road segments (Fig. 4e) for visualization and further processing. Road segments attributed with a high built-up area fraction presumably are located in areas that have been densely built-up in the year $T$, and thus, are likely to have existed at that point in time. Thus, we assume that such road segments are depicted in a given historical map of the year $T$. We will compare these fractions to the results of the road overlap detection in Section 3.3.

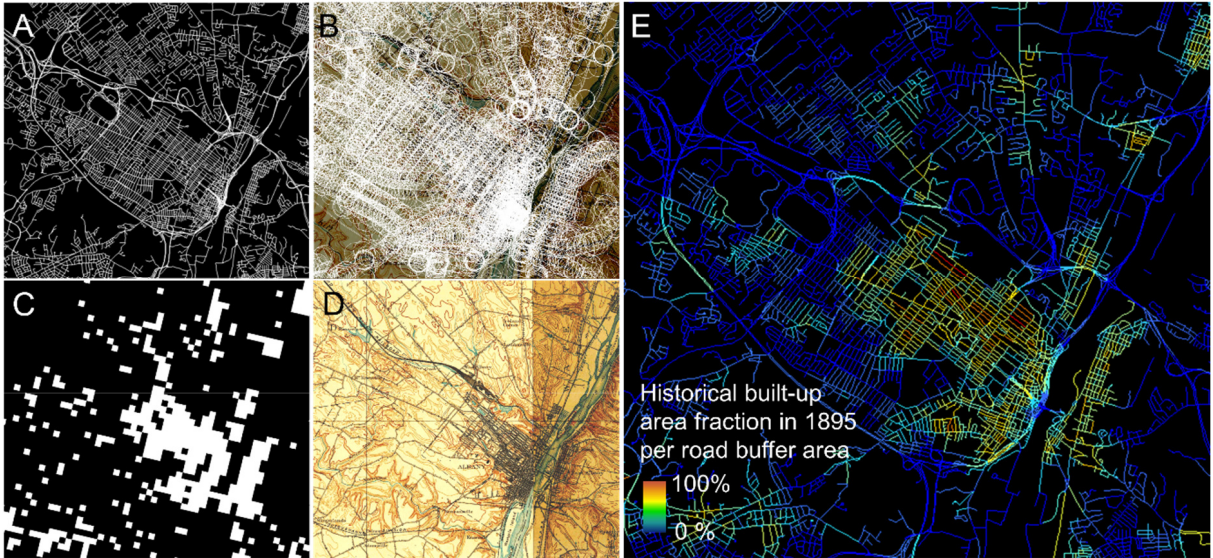

**Figure 4. Illustrating the automated reference data creation. (a) Contemporary road network, (b) road segments buffered by 125m, (c) historical built-up areas from the HISDAC-US, for 1895, and (d) historical map from the same area from 1897. (e) shows the HISDAC-US built-up area fractions in 1895 within the buffer areas shown in (b), attributed to the original road segments shown in (a), with low built-up area fractions in blue, and high built-up area fractions in yellow to red.**

### 2.4. Historical road network extraction

We extract the historical road network from the pool of contemporarily existing road segments using a pipeline that involves spatial data processing, image processing, and data analytics. More specifically, this pipeline involves four steps: (a) spatial data processing to generate cross-sectional sampling locations (Section 2.4.1, Fig. 5a), (b) vector-raster data integration to harvest color information from the historical maps at the sampling locations (Section 2.4.2, Fig. 5b), (c) image processing to generate a continuous road overlap indicator metric (Section 2.4.3, Fig. 5c), and (d) discretization of the continuous metric in order to identify the approximate historical road network (Section 2.4.4, Fig. 5d). This road overlap indicator is based on the assumption that if a given road segment existed in the year $T$, a historical map from year $T$ will contain a linear symbol that spatially coincides with the road axis from the contemporary road network vector data. Due to positional inaccuracies in the historical map (as a result of inaccurate georeferencing, paper map distortions, or other sources of uncertainty, see Uhl et al. 2018), the linear road symbol may also run in parallel to the road axis. The method described in this section aims to detect such linear symbols in the historical maps.

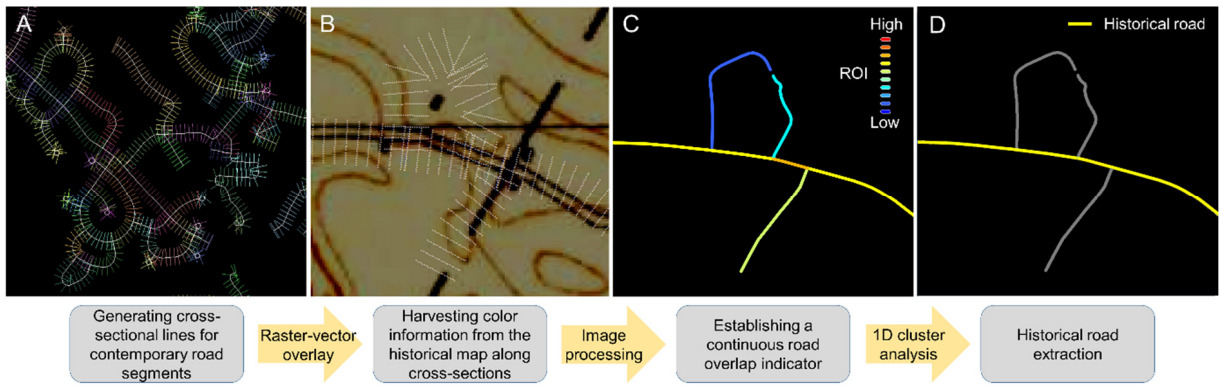

**Figure 5. Workflow for the historical road network extraction. (a) Contemporary road network segments with cross-sectional sampling lines, (b) cross-sectional sampling locations, (c) road-overlap-indicator (ROI) attached to the road segment features, and (d) identified historical roads after discretizing the ROI into two classes using 1D cluster analysis.**

### 2.4.1. Spatial data processing

For each map sheet comprising each study area – year combination, we project the contemporary road network data into the spatial reference system of each individual map sheet and clip the road vectors to the extent of the historical map sheet. We then generate cross-sectional sampling locations, which are arranged in lines perpendicular to the axis of each road segment (Fig. 5a,b). We access the vertices of each road vector object and calculate points along the road axis in a regular distance every 25m (=cross-section distance, *CSD*). At each of these locations, we generate a cross-section (i.e., a line perpendicular to the road axis) of length *CSL* (cross-section length) = 100m (50m in each direction). Examples of these cross-sectional lines are shown in Fig. 5a,b. The short distance *CSD* between cross-sections will allow to capture dashed lines, while the cross-section length *CSL* will allow to mitigate issues of offsets between road vector and road symbol in the map. Based on the spatial resolution of the historical map sheet underlying the road vector data (approximately 5m for the 1:62,000 scale maps used herein), we place 20 regularly spaced sampling points along each cross-sectional line (see Fig. 5b), resulting in a distance of 5m between sampling points, consistent with the pixel size of the scanned map images.

### 2.4.2. Vector-raster data integration

For each scanned and georeferenced map sheet of each study area and year, we automatically obtain the warping parameters (i.e., the affine transformation parameters to convert world coordinates into pixel coordinates). Based on these transformation parameters, we loop over the sampling points of the cross-sections of each road segment, identify the pixel of the underlying historical map, and register the R,G,B color information found for that pixel. This way, we generate a list containing the map color information for each map sheet, for each point in time, for all sampling locations of each cross-section, for each road segment.

### 2.4.3. Image processing and road overlap indicator calculation

Based on the systematically extracted color information in Section 2.4.2, we stack the RGB values for each cross-section horizontally, and then stack these horizontal lines vertically in their order of appearance along the road axis. Thus, for each road segment, we generate a pseudo-spatial axial image, with the y-direction corresponding to the road axis, and the x-axis corresponding to the directions perpendicular to the road axis. This way, we arrange the color information collected along a road axis of arbitrary shape (which may be curved) in a Cartesian, two-dimensional space (see Fig. 6, left column for some examples).

While the width *w* of these axial images is given by the number of cross-sectional sampling locations (i.e., 20), the height *h* is defined by the number of cross-sections and thus, is a direct function of the road segment length. For simplified data processing, we regularize these axial images into a common target shape of 20 × 20 pixels. This regularization consists of (a) random sampling (N=20) the rows of the image if h > 20; this occurs if road segments are longer than 20*25m=500m (Fig. 6a). Conversely, if h < 20, the road segment is short and we transform the image into the target shape, by using a reflection padding strategy to impute the missing values in the target grid of 20 × 20 pixels (Fig. 6b). In addition to that, we convert the RGB information into grayscale. The second column from the left in Fig. 6 shows a few examples of the regularized, grayscale axial images.

If a road from the pool of contemporary vector roads existed in a given historical map, we assume spatial coincidence or that road vector axis and road symbol run in parallel. Thus, we calculate the west-east image gradients within each axial image. As shown in Fig. 6 (third column from the left), these gradient maps are sensitive to the existence of a linear symbol in the map, parallel to the road vector axis. In order to quantify this sensitivity, we plot the north-south sums of the west-east gradients for each column and calculate the area under this aggregated gradient curve (Fig. 6, right column). This area under the curve is our road overlap indicator metric, which we call Road Overlap Indicator (ROI). As can be seen in the negative example (no road symbol in map, Fig. 6d), the ROI is expected to be low if no parallel linear feature exists in the historical map, and high, if otherwise. Moreover, the peak in the North-south sum curve indicates where the linear feature is located relative to the road axis. Here, it is worth noting that the magnitude of the ROI depends on the axial image dimensions, given by the chosen target shape, as well as on the contrast level in the axial image. Hence, the ROI is a metric that is directly comparable for road segments within but not across map sheets. This method is expected to be sensitive to any linear map symbol, parallel or coinciding with a contemporary road, such as railroads or contour lines (see Fig. 6e), which may result in misclassifications (see Section 3).

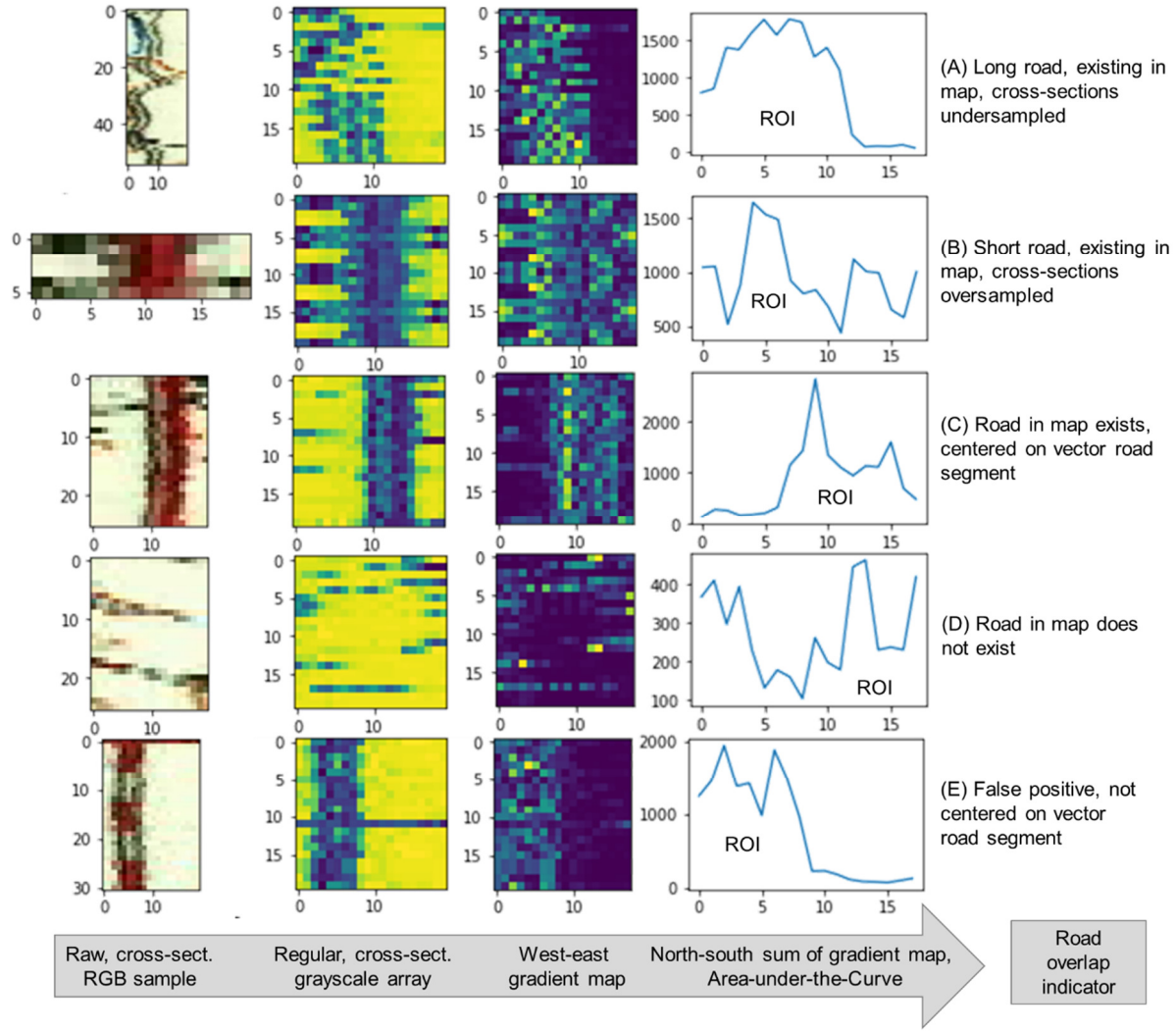

**Figure 6. Image processing steps to generate the road-overlay-indicator (ROI) measure, shown for five different scenarios. The ROI is the area-under-the-curve of the north-south sum of the west-east gradient map.**

### 2.4.4. Extraction of the historical road network through discretization

While the continuous ROI can be interpreted as a measure of likelihood that a road segment existed in a given historical map, many applications require binary estimates (i.e., road existed vs. road did not exist). Thus, in this last step, we discretize the continuous ROI into two classes, using ck-means (Wang & Song 2011) clustering. Ck-means is a variant of the k-means clustering algorithm, tailored to one-dimensional clustering problems. We preferred ck-means over other methods, as it allows to specify the desired number of clusters (i.e., 2), and it does not require the specification of data-specific parameters, nor does it make assumptions about density variations in the data, as opposed to methods such as DBScan (Schubert et al. 2017). Moreover, as ck-means has been developed for 1-dimensional data, it makes use of sorting functions and is expected to be highly performant.

After this clustering step, we calculate the average ROI per cluster, and assume that the cluster with higher average ROI represents the cluster of historical roads (i.e., that exist in the historical map). As previously discussed, the magnitude of the ROI depends on the predominant contrast level of a scanned map sheet, and thus, ck-means clustering is conducted separately for the road segments within each map sheet and compared to a "global" clustering strategy (i.e., across all map sheets of a study area – year combination.

### 2.4.5. Sensitivity analysis

The ROI measure is based on color information collected along cross-sectional lines perpendicular to the road center lines (Figure 5a,b), and derived from axial images constructed from the cross-sectional color information (Figure 6). The geometric properties of the cross-sections (i.e., cross-section length $CSL$ and distance between cross-sections $CSD$) and the dimensions of the axial windows (i.e., width $w$ and height $h$) potentially affect the magnitude of the ROI metric and thus, may affect the extracted road networks. For a subset of the NY study area (i.e., the 1895 map for the city of Amsterdam, NY) we systematically varied these four parameters and visually compared the ROI and the resulting clustering results for a range of scenarios (see Section 3.6).

## 2.5. Validation, cross-comparison, and plausibility checks

We carried out different types of diagnostic analyses to evaluate the quality of the ROI and of the extracted, historical road networks. These diagnostics include (a) visual assessments, (b) quantitative comparison against the two sets of reference data, including manually labelled reference data, and automatically created reference data based on historical building distributions from the HISDAC-US (see Section 2.3), as well as temporal plausibility checks.

### 2.5.1. Visual assessment

We visualized the ROI across two domains: 1.) in geographic space at the road segment level; and 2.) at the image patch level. For the second assessment, we classified the road segments into deciles based on the ROI distributions within each study area and year. We drew a random sample of N=9 segments per decile class, per study area, and year, and extracted the historical map content within a patch of size 500 m × 500 m around the segment centroid. We arranged the extracted map patches for visual assessment (Fig. 11). The results of the visual assessment are shown in Section 3.1.

### 2.5.2. Validation against manually annotated reference data

We analyzed the ROI distributions and conducted ROC analysis based on the manually annotated reference data, using the annotations (road presence/absence) as a binary variable, and the ROI as a continuous variable. In order to evaluate the quality of the extracted historical road network (i.e., after discretizing the ROI using ck-means, see Section 2.2.4), we report accuracy metrics such as precision, recall, and F1 score for different clustering scenarios. We report both instance-based accuracy metrics (i.e., based on the number of road segments in each agreement class: true positives $TP$, false positives $FP$, and false negatives $FN$), as well as road-length based accuracy metrics (see Heipke et al. 1997), which weighs the confusion matrices by the total length $L$ of road segments in each agreement class (Table 2). Such length-based accuracy metrics account for the irregular length of the segments which may bias the instance-based accuracy metrics. Moreover, such length-based accuracy metrics give a more realistic estimate of the uncertainty propagated into road length statistics derived from the extracted network. For example, if a few short roads are misclassified, their bias effect on the total historical road length is smaller than it would be for long, misclassified road segments (Section 3.3).

**Table 2. Instance and length-based accuracy metrics.**

|  | Precision | Recall | F1-score |
|---|---|---|---|
| Instance-based | $P_i = \dfrac{TP}{(TP + FP)}$ | $R_i = \dfrac{TP}{(TP + FN)}$ | $F1_i = 2\dfrac{P_i \cdot R_i}{(P_i + R_i)}$ |
| Length-based | $P_L = \dfrac{\sum L_{TP}}{(\sum L_{TP} + \sum L_{FP})}$ | $R_L = \dfrac{\sum L_{TP}}{(\sum L_{TP} + \sum L_{FN})}$ | $F1_L = 2\dfrac{P_i \cdot R_i}{(P_i + R_i)}$ |

### 2.5.3. Comparison to historical built-up areas from HISDAC-US

Based on the strategy described in Section 2.3.2, we annotated each of the 300,000+ road segments with the proportion of built-up area in the proximity of the road, as modeled by a buffer polygon around each road segment. This strategy yields a much larger sample than the manually collected reference data (Section 2.3.1) and thus, allows for more robust accuracy quantification. As discussed, we assume the presence of buildings to be indicative for the presence of a road. However, it is unknown what proportion of land in proximity of a road needs to be built-up to be confident about the presence of a road and thus, the choice of a specific threshold is difficult. Hence, we define a range of thresholds applied to the built-up area fraction associated with each road segment to create different sets of binary variables. We then compare these binary variables to the continuous ROI, using Receiver-operator-characteristic (ROC) analysis (Green & Swets 1966). ROC analysis is commonly used to evaluate the agreement in binary classification problems between a binary reference variable and continuous probability scores when the optimum threshold (that maximizes the true positive rate while minimizing the false positive rate) to be applied to the continuous variable is unknown. Here, we assume there is an ROI threshold that maximizes the binary agreement to the reference labels derived from the built-up area fractions. We conduct ROC analysis for each study area and year, and for a range of thresholds applied to the built-up area fractions, and visualize these ROC curves. Moreover, we analyze the distributions of the area-under-the-curve (AUC) (Fawcett 2006), of the maximum F1 score ($F1_{MAX}$), and of the ROI threshold associated with the $F1_{MAX}$. The ROC analysis results are presented in Section 3.2.

### 2.5.4. Temporal plausibility analysis

Finally, we assess the plausibility of our results over time. Here, we assume that road networks grow (i.e., expand or densify) over time. Thus, a road detected in a map of year $T$ needs to be detected in a later year $T+x$ as well. We test this hypothesis by visually comparing the ROI for a given road segment in $T$ and $T+x$, and by calculating the change in total road network length over time, which is assumed to be positive. Moreover, we compare the extracted road networks of subsequent years in a binary fashion, by calculating the transitions of a given road segment over time (e.g., road detected in $T$, but not in $T+x$) (Section 3.4).

Figure 7 summarizes the datasets and data processing steps used for the historical road extraction (left part), and the analytical steps performed for the different validation efforts (right part).

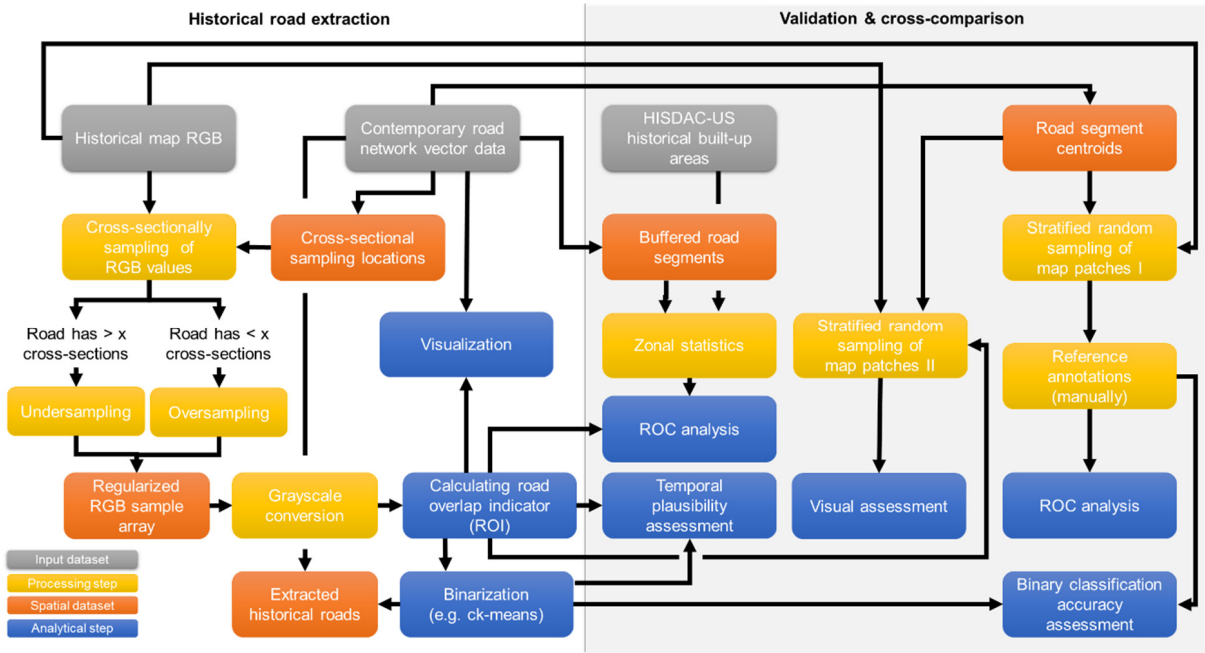

**Figure 7. Datasets, data processing and analytical steps used in this work. The steps performed to extract historical road networks are shown to the left, and the validation and cross-comparison is shown to the right. Colors represent the input datasets (grey), data processing steps (yellow), intermediate and output spatial datasets (red), and analytical steps (blue).**

## 2.6. Data processing and analysis tools

We collected HTMC historical maps using Python 3.7, and used GDAL/OGR[1] for automated map collar removal. We used the ESRI ArcPy[2] Python package to preprocess the NTD road network vector data, and GDAL/OGR Python package to generate cross-sectional sampling locations. We then used ESRI ArcPy for vector-raster data integration, and NumPy[3] Python package for image processing and establishing the road overlap indicator metric. Cluster analysis was done using a Python ck-means implementation[4], patch-based validation data was extracted using GeoPandas[5] and OpenCV[6] Python packages. The evaluation and validation experiments were conducted using Scikit-learn[7] Python package. Data visualization was done in ESRI ArcMap 10.8[8], as well as using Matplotlib[9], GeoPandas, and Seaborn[10] Python packages.

## 3. Results

In this section, we present the different results of the analyses conducted herein. First, we carried out a visual assessment: We map the ROI associated with each road segment in geographic space, and visualize a random sample of map patches collected at road segments in a continuum of the ROI obtained for each segment (Section 3.1). Second, we present the ROC analysis results from comparing the ROI against manually collected reference labels (i.e., road presence / absence) at the map patch level (Section 3.2). Third, we discuss the ROC analysis results from comparing the ROI against the built-up area fractions calculated from the HISDAC-US for each road segment (Section 3.3). Forth, we compare the extracted historical road networks (i.e., after binarizing the continuous ROI into two classes) against both, the manually collected, patch-level reference labels and the automatically created built-up area fractions (Section 3.4). Finally, we assess the plausibility of our results over time, in three analytical parts: (a) by comparing the ROI for a given road segment over time, (b) by assessing road network growth over time, and (c) by tracking the binary classes assigned to the road segments (i.e., historical versus more recent road) over time (Section 3.5).

---

[1] https://gdal.org/api/python.html
[2] https://pro.arcgis.com/en/pro-app/latest/arcpy
[3] https://doi.org/10.1038/s41586-020-2649-2
[4] https://github.com/llimllib/ckmeans
[5] https://doi.org/10.5281/zenodo.3946761
[6] https://docs.opencv.org
[7] https://scikit-learn.org/
[8] https://www.esri.com/en-us/arcgis/products/arcgis-desktop/resources
[9] https://matplotlib.org/
[10] https://seaborn.pydata.org/

### 3.1. ROI visualization

We calculated the road overlap indicator (ROI) metric for each of the 60,000+ road segments, for all points in time, and visualized the ROI attributed to each road segment of the contemporary road network. We expect regions of high ROI values in the historical centers of dense, urban areas, that grow over time. Across all three study areas, the results are largely plausible, i.e., we observe high ROI values in densely settled urban centers, and these areas of high ROI levels generally expand over time (Fig. 8, Fig. 9). Visual comparison of the ROI values for roads in the historical Albany city center (Fig. 8c,d) indicates high levels of sensitivity of the ROI, in particular in dense urban areas, and the magnitude of the ROI in these urban areas appears to increase over time. Moreover, we observe that most rural roads in the Greater Albany area (Fig. 8a) already existed in the 1890s, whereas most roads in dense, urban roads did not exist yet by the 1950s.

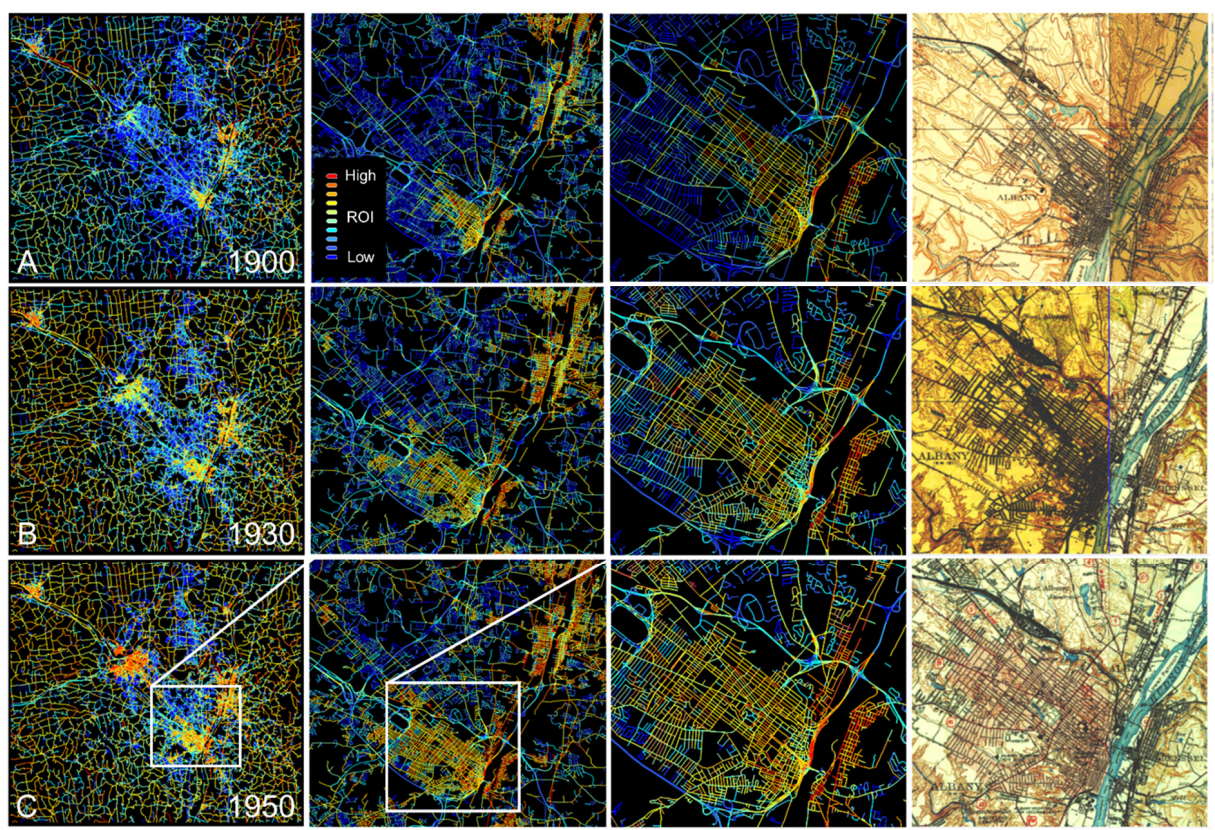

**Figure 8. Maps showing the road overlap indicator (ROI) metric for the Albany (NY) study area in (a) 1900, (b) 1930, and (c) 1950, at different level of detail. Shown is the ROI for each road segment, indicating whether a road likely existed in the year the historical map was created (red, yellow), blue roads likely did not exist yet. The column to the right shows the underlying historical maps for comparison.**

Similar observations can be made for the California study area (Fig. 9) where the spatial distributions of the ROI illustrate the overall growth patterns in the Bay area (Fig. 9a). In the Santa Clara enlargements (Fig. 9b), we observe that the ROI magnitude appears to decrease over time, which is due to the higher levels of contrast in the 1890 maps as compared to the 1950 maps where urban road networks are depicted based on pink colors (Fig. 9b,c). Worth noting are also the large amounts of contemporary roads with low ROI, neither existing in the 1890s nor the 1950s. These patterns impressively illustrate the extreme urban growth that the Bay area has experienced since the 1950s (Fig.9b).

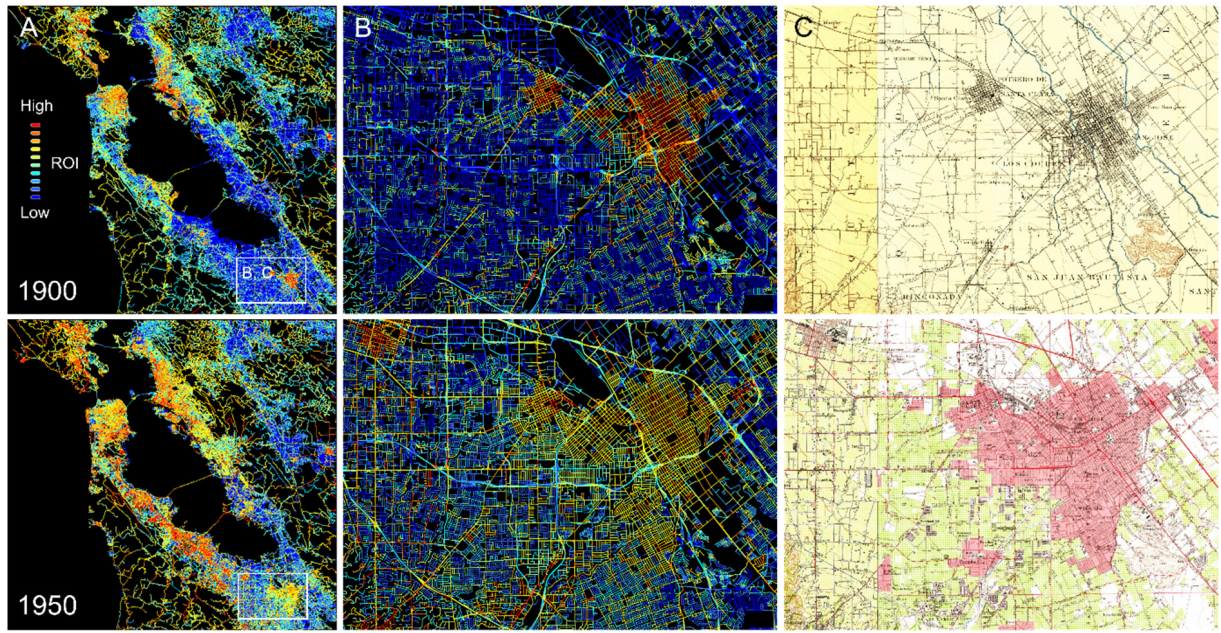

**Figure 9. Maps of the road overlap indicator (ROI) metric for the Bay area (California) study area in 1900 and 1950, shown (a) for the whole study area, and (b) for Santa Clara, California. Shown is the ROI for each road segment, indicating whether a road likely existed in the year the historical moa was created (red, yellow), blue roads likely did not exist yet. Panel (c) shows the historical maps underlying the ROI metrics in (b).**

The Mobile study area represents the most challenging study area, as the map is a rather coarse, unspecific black-and-white print. Here we observe a slightly different picture (Fig. 10). While large parts of the contemporary road network, attributed with low ROI values, did not exist in the 1920s, we observe that also for maps of this cartographic style, the ROI is sensitive to the dense urban street network that coincides or runs in parallel to the contemporary road network data (Fig. 10b,c). However, the variation of the ROI across the Mobile study area is much less than what we observed in the Albany and California study areas, likely due to the high contrast levels across all Mobile map images. Note that the ROI color scaling is consistent across Figures 8, 9, and 10.

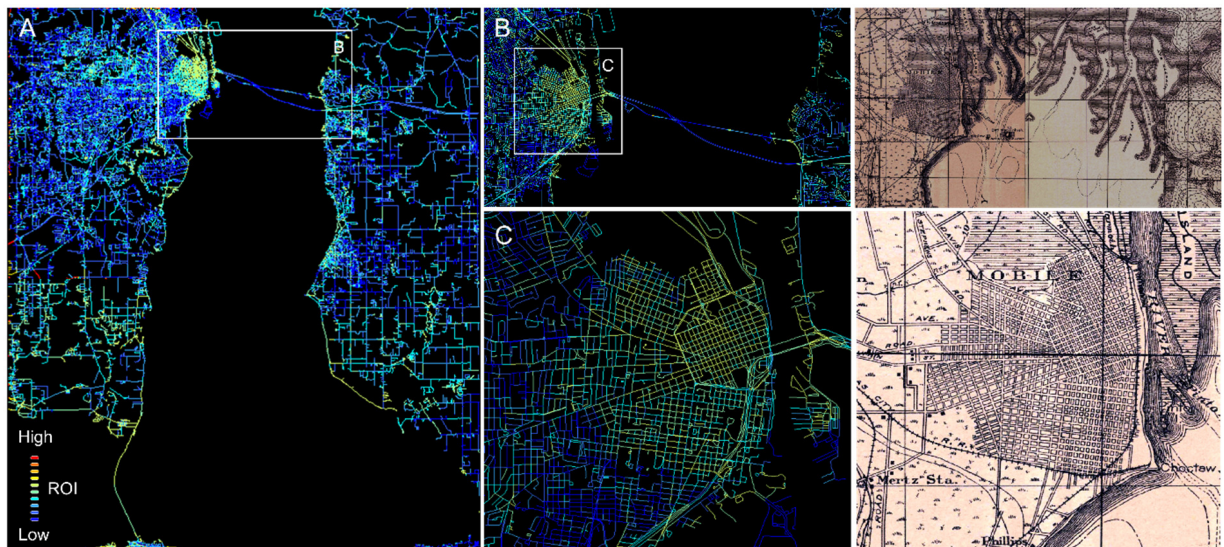

**Figure 10. Maps of the road overlap indicator (ROI) metric for the Mobile Bay (Alabama) study area in 1920, (a) shown for the whole study area, and (b)-(c) shown for the city of Mobile, at different levels of detail, including the underlying historical map. Shown is the ROI for each road segment, indicating whether a**

**road likely existed in the year the historical map was created (light blue, yellow), dark blue roads likely did not exist yet.**

Such visualizations across geographic space exhibit high levels of plausibility, and the visualization of map patches cropped around a stratified sample of road segments (see Section 2.5.1), sorted by the magnitude of the ROI obtained for each segment show a similar picture. We expect to see very few road symbols in map patches assigned with low ROI values, and at least one road symbol in map patches associated with high ROI values. As shown in Fig. 11, the frequency of map patches containing one or more road symbols increases with increasing ROI, and the highest ROI values are found either for dense urban road networks, or for individual rural road symbols without other map content.

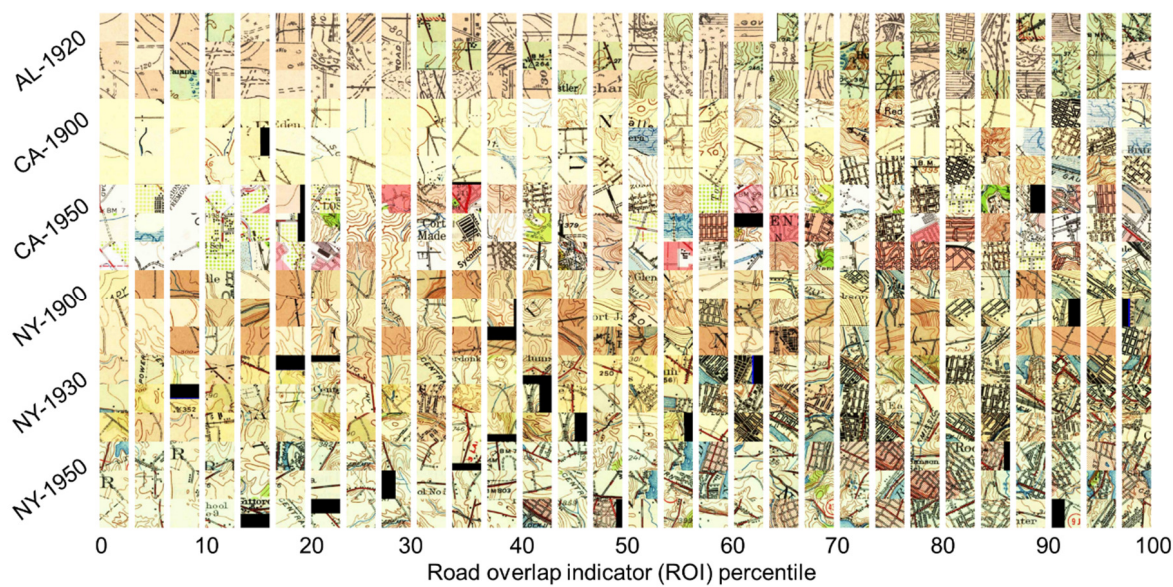

**Figure 11. Visual assessment of the road overlap indicator metric (ROI) at the patch level, by arranging a random sample of map patches collected at road segment centroids, by the ROI associated with each road segment. Map patches are sorted from left to right by their ROI.**

### 3.2. ROI evaluation against map-based reference data

The comparison of the ROI assigned to the road segments and the manually created presence / absence labels for a subset of approximately 1,200 road segments exhibits interesting, quantitative insight of the sensitivity of the ROI metric to the presence of road symbols in historical maps. We expect high ROI values for road segments attributed with "road presence" in the reference labels, and vice-versa, and a clearly defined ROI threshold that achieves high levels of agreement to the reference labels. The distributions of the ROI for each group of reference labels are shown in Fig. 12a, overall and for rural and urban strata. Higher values of the ROI are encountered if roads are present, and low values are found if road symbols are absent in the map, and this pattern appears to be more pronounced in urban areas. The ROC plots for individual study areas (Fig. 12b) indicate high levels of variability of these associations across study areas, with a clear progression of increasing AUC values over time in both, the NY study area (AUC = 0.77, 0.86, 0.96) and for the CA study area (AUC = 0.70, 0.76). The lowest AUC value is found for the AL study area, which is expected due to the simplistic and coarse cartographic style used in the Mobile map. Fig. 12c shows the F1-score for each ROI threshold used for binarization. While the magnitudes of the F1-score roughly reflect the patterns observed in the AUC values, it is notable that the ROI threshold that maximizes the agreement with the reference labels ranges for most study areas between 8,000 and 11,000; however, there are no nuanced peaks. This indicates that a simple and generally applicable threshold on the continuous ROI for extracting the historical road network at acceptable levels of accuracy does not exist.

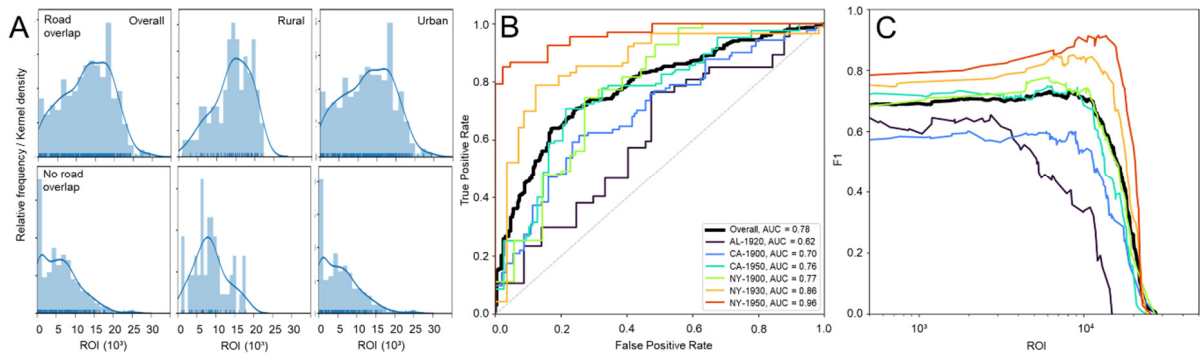

**Figure 12.** Evaluation of the ROI against manually created reference data. (a) ROI distributions, stratified by manually created reference labels, overall and in rural/urban strata. (b) ROC curves of the ROI metric and manually created reference labels overall, and for each study area, and (c) corresponding F1-scores plotted against the separation thresholds.

### 3.3. Comparing the ROI to historical built-up areas

While the evaluation results against manually labelled reference data show promising and plausible trends, the statistical support (N=1,200) is relatively low, compared to the building-based reference data, available for most of the 60,000 road segments (see Section 2.3.2). These reference data consist of the fraction of built-up area in proximity of each road segment. Under the assumption that the co-evolution of roads and buildings is largely coherent, we expect high levels of ROI where the historical built-up area fraction is high. Thus, for a given threshold applied to the built-up area fractions, a ROC analysis against the ROI values should yield relatively high levels of the AUC measure. Moreover, we test whether there is a threshold that can be applied to the ROI in order to maximize the agreement between binarized distributions of the ROI and built-up area fractions.

In analogy to the ROC analysis conducted against the manually labelled reference data, we present the ROC plots of the ROI against building presence-absence labels in Fig. 13. We tested different thresholds applied to the built-up area fraction in each road buffer (see Section 2.3.2) and observed that the AUC values are highest for thresholds of 75% and 90%, i.e., the ROI separates well the roads in areas of high built-up density from those in areas of lower built-up density. These increasing AUC trends also persist in most urban and rural strata (Appendix Fig. A2). The AUC values obtained from the ROC analysis across each study area are slightly lower than the AUC values obtained from comparison to manually (map-based) reference data (Section 3.2). However, we would like to reiterate that these results need to be interpreted carefully, as we use the presence of buildings to evaluate the presence of roads. Thus, these results are directly affected by the relationship between building and road presence, and we assume that the co-evolution of roads and buildings is largely coherent (Achibet et al. 2014). However, variations in this relationship across study areas and time periods may exist. To test this, we cross-compared the map-based, manually created (and highly reliable) reference data against the continuous built-up area fractions measured along the roads (see Section 2.5.3). As shown in Fig. A3, the relationship of built-up fraction and road presence is ambiguous, in both urban and rural regions (Fig. A3,a), and the AUC values obtained from the ROC analysis vary strongly across study areas (Fig. A3b). Moreover, the maximum agreement (as measured by the F1-score) is achieved if we use a threshold of >0% built-up area fraction. The latter observation implies that if there is one building mapped in a historical map, the likelihood of the presence of a road in direct vicinity is very high. While this threshold of >0% contradicts the threshold of >75%, for which the AUC of the ROI against the building-based reference labels maximizes, this discrepancy needs to be understood as a direct consequence of using two reference data types of different nature.

Furthermore, the ROC curves shown in Fig. 13 exhibit high levels of variability across individual map sheets (grey lines), indicating (a) high levels of variability of the ROI across map sheets, or (b) high levels of variability of the co-evolution of roads and buildings across map sheets. Notably, the map sheets in the Alabama study area exhibit very nuanced peaks for an optimum separation of historical versus more recent roads at the map sheet level (Fig. 13f). This is likely due to the homogeneous map styles used in this study area.

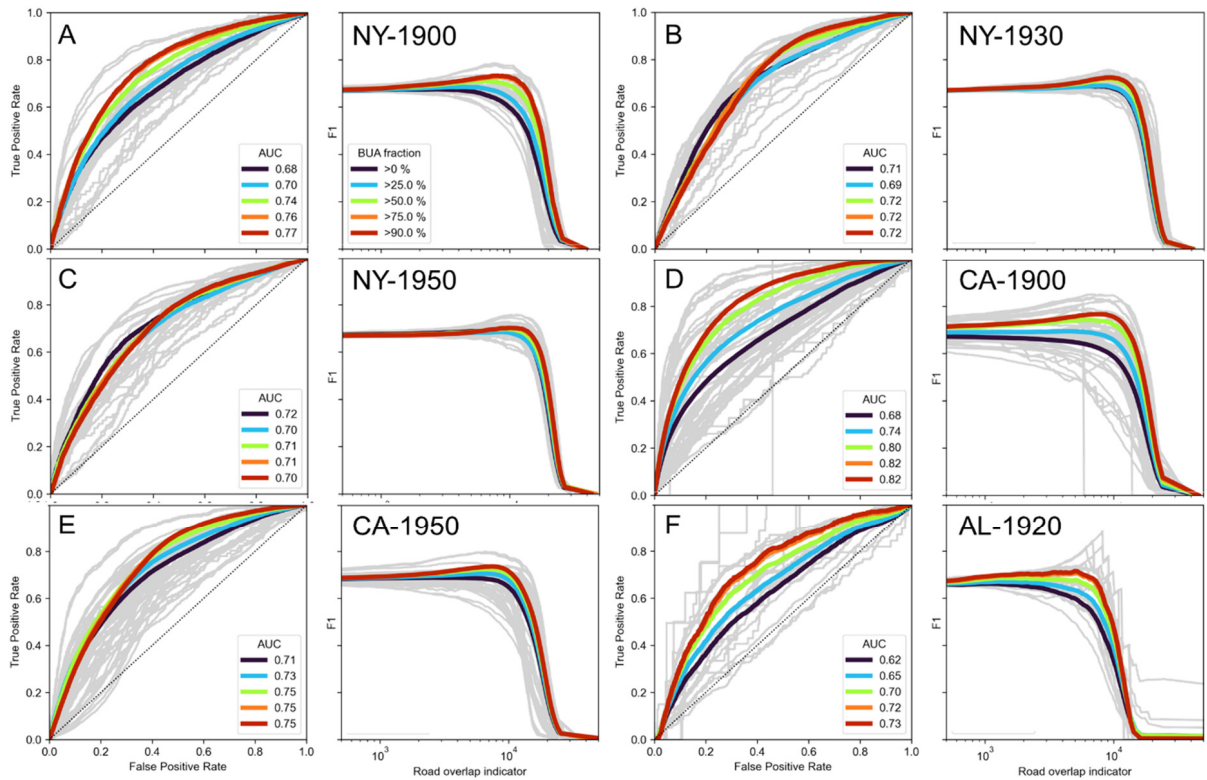

**Figure 13.** ROC analysis results against building-based reference data for different thresholds of built-up area fraction: ROC plots and optimum ROI thresholds (based on the maximum F1-score) for the NY study area in (a) 1900, (b) 1930, (c) 1950, for the CA study area in (d) 1900, (e) 1950, and (f) for the AL study area in 1920, Grey lines show the ROC curves for each individual map sheet (grey) per study area.

This variability across map sheets is also observed in Fig. 14, that shows the distributions of AUC, maximum F1-score, and the optimum ROI threshold obtained at the map sheet level within each study area, shown only for the built-up fraction thresholds >75% and >90%, for which maximum AUC values are achieved in Fig. 13. These distributions are shown separately for the rural and urban stratum, and indicate that the agreement of the ROI with built-up area derived road presence labels is higher in urban areas than in rural areas (i.e. higher levels of AUC and maximum F1-score, Fig. 14a,b). Moreover, we observe high variation of the optimum ROI threshold, both across study areas and rural-urban strata (Fig. 14c). This is in line with the evaluation against manually created reference data (Section 3.2) and further indicates that a binary separation of roads that exist in a given historical map based on a simple, global threshold is not feasible.

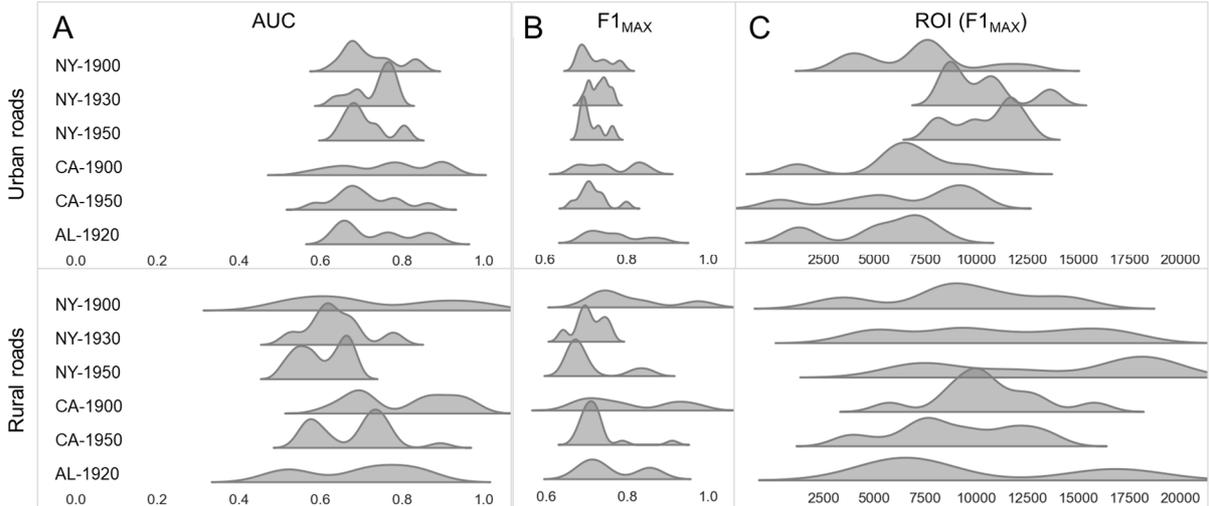

**Figure 14.** Distributions of AUC, maximum F-1 score, and ROI threshold corresponding to the maximum F-1 score, within each study area, and per road length stratum (roughly separating short urban from long rural roads), shown for the 0.75 and 0.9 built-up area fraction thresholds only.

### 3.4. Evaluation of historical road networks extracted by ck-means binarization

The evaluation of the continuous ROI against both, manually created and built-up area based reference labels (Sections 3.2, 3.3), indicates there is no global and generally applicable threshold that can be applied to the continuous ROI measure in order to create an accurate distinction between historical and more recent roads. Based on our results, a data-driven approach to discretize the ROI into two classes seems to be a suitable solution. Moreover, we observed strong variation of the evaluation results per map sheet, which implies that such a data-driven approach needs to be applied to the map-sheet level distributions of the ROI. In this subsection, we show the binarized clustering results from applying ck-means to the ROI distributions, representing the extracted historical road networks, and evaluate the agreement of these binary labels (i.e., historical versus more recent road) with both the manually created, and automatically created, building-based reference labels.

Figure 15a shows the extracted historical road networks for the three study areas and all years (in yellow), after applying ck-means partitioning to the *map-sheet level* ROI distributions (see Section 2.4.4). For comparison, we also conducted ck-means partitioning based on the *study-area level* ROI distributions, and show the agreement and disagreement between the two clustering strategies in the agreement-disagreement maps in Fig. 15b. Notably, the different clustering strategy yield similar results for the NY study area, where map sheets are of similar content and level of contrast, but cause some differences in the Northwest of the CA study area (characterized by dense contour lines), and in the Southwest of the AL study area (hatched wetland symbols).

Furthermore, we calculated Silhouette scores (Rousseeuw 1987) for each road segment, based on the clusters assigned by the ck-means method. The silhouette scores indicate the proximity of a data point to the cluster center and indicate the "confidence" of a data point being a member of the assigned cluster. As shown in Appendix Fig. A4, negative (i.e., more recent) road segments often exhibit higher silhouette scores than the historical segments, which may be used as additional information for the interpretation of the results shown in Fig. 14. For example, road segments that "switch" cluster membership when applying map-sheet level versus study-area level clustering (e.g., Southwest of the AL study area, Fig. 15b) exhibit lower Silhouette scores, indicating lower levels of classification confidence.

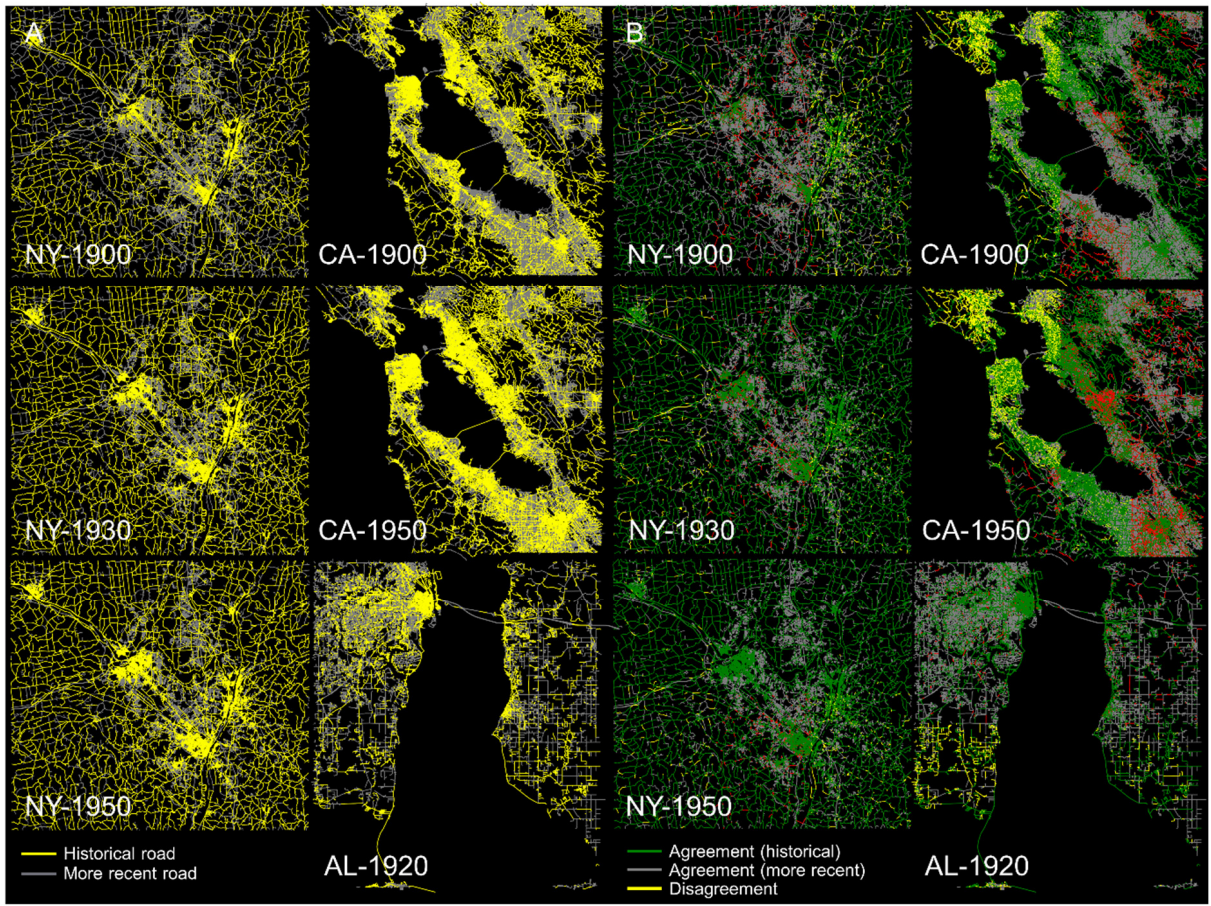

**Fig. 15.** Extracted historical road networks for the three study areas and the different years. (a) map-sheet level ck-means clustering, and (b) agreement-disagreement map of map-sheet level cluster analysis versus study-area level cluster analysis.

Moreover, we quantitatively compared the extracted historical road networks against both reference datasets, the manually created, map-based reference labels, and the automatically created reference labels based on the road buffer built-up fraction. As we observe highest agreement (i.e., F1-score) between the two types of reference data for a built-up fraction >0% (Fig. A4c), we report the agreement of the historical road networks against the built-up-fraction based reference data for that threshold only (Table 3). As can be seen in Table 3, the agreement appears to be higher with map-based reference data than with built-up-fraction based reference data. This is expected due to potential variations in the relationship between the evolution of buildings and roads.

Also, additional uncertainty in the building-based reference labels may arise from overlapping buffer areas in dense, urban road networks (Fig. 4b). Generally, accuracies are highest in the NY study area, with increasing trends from the 1893 epoch to the 1950 epoch. As observed previously, accuracies are lowest in the Mobile study area, likely due to the simplistic cartographic design and the large amounts of other linear features such as wetland symbols, etc. Moreover, we observe variations of accuracy between the urban and rural strata in most study areas (Appendix Table A1). These urban-rural variations do not seem to follow a clear trend, which may be attributed to the higher levels of uncertainty in the building-based reference data, as previously discussed. Interestingly, the effect of the clustering strategy (per study area or per map sheet) only has a minor effect on the accuracies, without exhibiting a clear trend. Hence, these results do not allow for specifying recommendations on which of the strategy to use.

**Table 3. Binary classification accuracy assessment of the ck-means clusters against manually created reference data and reference labels based on the built-up area fraction per road buffer (Prec = Precision, Rec = Recall).**

| Referene data source: | | | reference labels from built-up fraction >0% | | | | | | manually created reference labels | | | | | |
|---|---|---|---|---|---|---|---|---|---|---|---|---|---|---|
| Cluster analysis setup: | | | per study area | | | per map sheet | | | per study area | | | per map sheet | | |
| Study area | Year | Weighting scheme | Prec | Rec | F1 | Prec | Rec | F1 | Prec | Rec | F1 | Prec | Rec | F1 |
| AL | 1920 |  | 0.12 | 0.43 | 0.19 | 0.17 | 0.43 | 0.24 | 0.67 | 0.50 | 0.58 | 0.69 | 0.36 | 0.48 |
| CA | 1900 |  | 0.42 | 0.73 | 0.53 | 0.40 | 0.57 | 0.47 | 0.52 | 0.68 | 0.59 | 0.58 | 0.62 | 0.60 |
| CA | 1950 | Road length based | 0.74 | 0.58 | 0.65 | 0.79 | 0.57 | 0.66 | 0.89 | 0.67 | 0.77 | 0.83 | 0.57 | 0.67 |
| NY | 1900 |  | 0.63 | 0.64 | 0.63 | 0.67 | 0.64 | 0.65 | 0.75 | 0.77 | 0.76 | 0.80 | 0.77 | 0.79 |
| NY | 1930 |  | 0.83 | 0.72 | 0.77 | 0.84 | 0.72 | 0.77 | 0.84 | 0.93 | 0.89 | 0.84 | 0.93 | 0.88 |
| NY | 1950 |  | 0.77 | 0.78 | 0.78 | 0.76 | 0.73 | 0.74 | 0.97 | 0.93 | 0.95 | 0.97 | 0.88 | 0.92 |
| AL | 1920 |  | 0.19 | 0.36 | 0.24 | 0.22 | 0.36 | 0.27 | 0.67 | 0.39 | 0.49 | 0.61 | 0.30 | 0.41 |
| CA | 1900 |  | 0.48 | 0.70 | 0.57 | 0.50 | 0.70 | 0.58 | 0.63 | 0.58 | 0.60 | 0.65 | 0.58 | 0.62 |
| CA | 1950 | Instance-based | 0.76 | 0.61 | 0.68 | 0.77 | 0.61 | 0.68 | 0.86 | 0.62 | 0.72 | 0.84 | 0.59 | 0.69 |
| NY | 1900 |  | 0.70 | 0.60 | 0.65 | 0.73 | 0.61 | 0.67 | 0.78 | 0.72 | 0.75 | 0.81 | 0.74 | 0.77 |
| NY | 1930 |  | 0.85 | 0.71 | 0.77 | 0.86 | 0.72 | 0.78 | 0.89 | 0.83 | 0.86 | 0.89 | 0.83 | 0.86 |
| NY | 1950 |  | 0.86 | 0.66 | 0.75 | 0.84 | 0.65 | 0.73 | 0.97 | 0.84 | 0.90 | 0.96 | 0.83 | 0.89 |

### 3.5. Temporal plausibility analysis

Lastly, we assess the plausibility of our results over time. As we assume road network growth to be predominant in our study areas (as opposed to road network shrinkage), we consider road networks that "disappear" over time as implausible and likely to be the result of misclassification. As described previously, this plausibility assessment consists of three analytical parts: (a) assessing the trend of the ROI assigned to a given road segment over time (Fig. 16), (b) assessing the total length of the extracted historical road networks over time (Table 4), and (c) tracking the binary classes assigned to the road segments (i.e., historical versus more recent road) over time (Table 5).

We begin with analyzing the continuous ROI metrics, which we group into 50 equal width classes and cross-tabulate them between subsequent points in time $T_1$ and $T_2$ for the same road network segment (Fig. 16). We observe that many value pairs are located near or above the main diagonal (i.e., $ROI_{T2} \geq ROI_{T1}$), indicating that these road segments either exist in both historical maps, or do not exist in either of the maps. Data points above the main diagonal correspond to road segment with an increasing ROI over time, representing roads that were newly built during the respective time periods. Only few road segments exhibit a ROI decrease over time (i.e., they are located below the main diagonal), which would indicate "disappearing" road symbols over time (e.g., for rural roads in NY 1930-1950). This qualitative, visual assessment indicates largely plausible temporal trajectories of the ROI over time.

After discretizing the continuous ROI into historical and more recent road segments (Section 2.4.4), we extracted road network statistics such as the number of segments and their corresponding road length, as well as the relative change in road length between the points in time $T_1$ and $T_2$ (in %, referred to the road network length in $T_1$). These statistics are shown in Table 4. Notably, historical road segments from each point in time were extracted independently. Nevertheless, we observe increases in the total road length over time, and mostly positive change rates, which seems plausible. Table 4 also shows these statistics for both clustering strategies (per map sheet, and per study area). The change rate in the NY study area from 1930 to 1950 is slightly negative for study-area level clustering, and switches to a positive change rate when applying map-sheet level clustering, potentially indicating that the map-sheet level clustering strategy yields more plausible results. Moreover, these network statistics over time do not only serve as a plausibility check, but also illustrate a potential application of the proposed approach, providing quantitative insight into the early evolution of road networks over long time periods of time, derived in a fully automated manner.

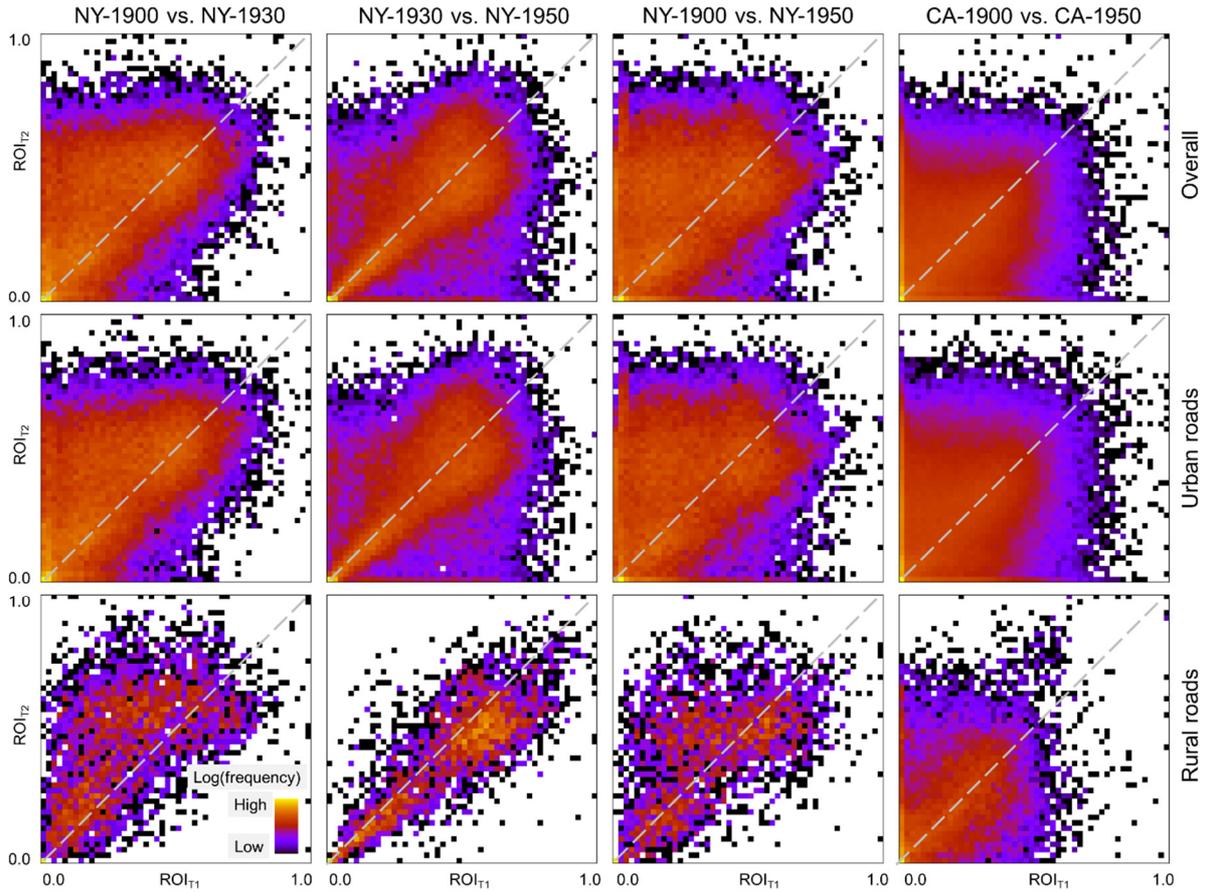

**Figure 16.** Temporal plausibility analysis assessing the trend of the ROI obtained for each road segment over time, for the NY and CA study area, and for all combinations of points in time. Shown are the bivariate histograms of ROI pairs after discretizing the ROI distributions into 50 equal-width classes. A concentration on or above the main diagonal implies roads that persist, or that were added during the observational window, respectively. Data points below the main diagonal indicate disappearing roads, which is considered implausible. High frequencies at or near (0,0) indicate roads that did not exist during the study period (i.e., low ROI in both points in time).

**Table 4.** Road length dynamics over time. The reported relative change is measured as the increase in road network length relative to the earlier point in time.

| | | Albany (NY) study area | | | | | | Bay Area (CA) study area | | |
|---|---|---|---|---|---|---|---|---|---|---|
| | | Total road network length [km] | | | Relative change [%] | | | Total road network length [km] | | Relative change [%] |
| | Year | 1900 | 1930 | 1950 | 1900-1930 | 1930-1950 | 1900-1950 | 1900 | 1950 | 1900-1950 |
| Clustering per study area | overall | 4,359 | 6,142 | 6,382 | 40.9 | 3.9 | 46.4 | 14,572 | 18,557 | 27.4 |
| | urban | 2,085 | 3,054 | 3,317 | 46.5 | 8.6 | 59.1 | 8,283 | 11,461 | 38.4 |
| | rural | 2,274 | 3,088 | 3,065 | 35.8 | -0.7 | 34.8 | 6,289 | 7,096 | 12.8 |
| Clustering per map sheet | overall | 4,181 | 6,077 | 6,292 | 45.3 | 3.5 | 50.5 | 14,143 | 18,774 | 32.7 |
| | urban | 2,054 | 3,087 | 3,300 | 50.3 | 6.9 | 60.7 | 8,017 | 11,503 | 43.5 |
| | rural | 2,127 | 2,990 | 2,992 | 40.6 | 0.1 | 40.7 | 6,126 | 7,271 | 18.7 |

Lastly, we assessed the changes over time at the road segment level with respect to the contemporarily existing road network, by cross tabulating the binary labels (i.e., historical versus more recent roads) for the same road segments in subsequent points in time (Table 5). These results provide some insight on the road network expansion over time (e.g., roughly a quarter of the contemporary road networks in the CA and NY study areas were built between 1900 and 1950). Moreover, we observe that the road segments that "disappeared" over time (i.e., labelled as "historical" in $T_1$, but not in $T_2$) which we
23

consider implausible, are consistently lower when using the map-sheet level clustering strategy, as compared to the study-area level clustering. This trend also persists in rural and urban strata (Appendix Table A2). While we did not observe such a trend in the accuracy analysis (possibly as a result of the small sample size not capturing these improvements), this observation confirms our initial hypothesis that a cluster analysis per map sheet is expected to improve the extraction results, as it reduces bias introduced by heterogeneous contrast levels across different map sheets.

**Table 5. Multi-temporal road network statistics based on cross-tabulation of cluster labels over time.**

| Clustering strategy | Study area | T1 | T2 | km road | | | | [%] of contemporary road length | | | |
|---|---|---|---|---|---|---|---|---|---|---|---|
| | | | | not existent | persistent | newly built | disappeared | not existent | persistent | newly built | disappeared |
| Clustering per study area | NY | 1900 | 1930 | 3,625 | 3,756 | 2,522 | 605 | 34.50 | 35.75 | 24.00 | 5.75 |
| | NY | 1930 | 1950 | 3,399 | 5,664 | 825 | 613 | 32.36 | 53.94 | 7.86 | 5.84 |
| | NY | 1900 | 1950 | 3,379 | 3,727 | 2,762 | 633 | 32.18 | 35.49 | 26.31 | 6.03 |
| | CA | 1900 | 1950 | 12,966 | 10,544 | 9,253 | 4,578 | 34.72 | 28.24 | 24.78 | 12.26 |
| Clustering per map sheet | NY | 1900 | 1930 | 3,317 | 3,772 | 2,277 | 503 | 33.61 | 38.22 | 23.07 | 5.10 |
| | NY | 1930 | 1950 | 3,105 | 5,572 | 716 | 477 | 31.46 | 56.45 | 7.25 | 4.84 |
| | NY | 1900 | 1950 | 3,082 | 3,775 | 2,513 | 501 | 31.22 | 38.25 | 25.46 | 5.08 |
| | CA | 1900 | 1950 | 13,026 | 10,142 | 8,097 | 4,096 | 36.84 | 28.68 | 22.90 | 11.58 |

### 3.6. Sensitivity analysis

Varying the parameters cross-section length *CSL* and cross-section distance *CSD* used to harvest color information from the historical maps (Fig. 5), as well as the dimension of the axial images (i.e., *w* and *h*, Fig. 6) used to construct the ROI metric potentially affect its magnitude and thus, may affect the extracted road networks. As shown in Fig. 17a, varying the dimension of the axial images, for fixed values of *CSL* = 100 m and *CSD* = 25 m, do affect the magnitude of the ROI, but result in highly similar spatial patterns and clustering results. When varying the parameters of the cross-sections we observe that for a very low cross-section length (*CSL*=25 m) the resulting clusters do not seem to capture the historical road network very well (Fig. 17b). This is likely due to positional offsets between contemporary road network vector data and the historical map exceeding 25 m. Choosing large values for *CSL* and / or *CSD* (Fig. 17c,d) does not have a major effect on the outcomes but will possibly affect processing time. Thus, the choice of a short cross-section length *CSL*, long enough to account for positional offsets, in combination with a high *CSD* may increase processing efficiency while keeping extraction quality constant when working with large volumes of data. See Fig. A5 for the full sensitivity analysis results varying all parameters systematically.

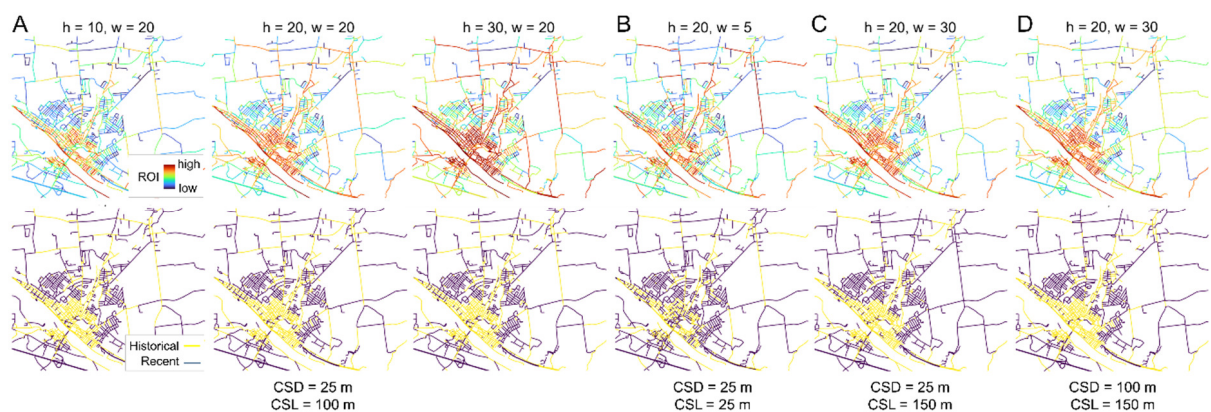

**Figure 17. Sensitivity analysis results. Road overlap indicator (ROI, top row) and resulting ck-means clusters (bottom row) based on a map of Amsterdam (NY) from 1895, shown for different combinations of**

**cross-section length (*CSL*) and cross-section distance (*CSD*) to collect color information from the historical map, and for varying dimensions (*h,w*) of the axial images used to construct the ROI. (a): Fixed *CSD* of 25 m and *CSL* of 100 m as used in this analysis, but for varying levels of *h* and *w*. (b) – (d): varying *CSD* and *CSL* for a fixed window height. Note that the window width *w* is linked to the *CSL* parameter and thus, varies as well.**

## 4. Discussion

The results presented in Section 3 illustrate that the proposed method to assign a road overlap indicator to each contemporary road network segment works and responds to the presence of linear road symbols in a given historical map (Fig. 11). Results are geographically logical (i.e., highest ROI values are observed in clusters, typically located in the center of contemporary cities and thus, likely to represent a spatial approximation of historical urban centers existing in the reference year of the historical map, including the roads connecting these centers (Figs. 8-10). Moreover, the temporal trends of the extracted road networks are plausible, i.e., they suggest road network growth over time (Tables 4,5, Fig. 16), which is in line with the literature on road network development (Levinson 2005). Finally, the proposed method appears to be insensitive to the choice of parameters used to sample color information from the historical maps and to construct the ROI measure (Fig. 17).

The extracted historical road networks reflect long-term urbanization and land development patterns and constitute an important step towards the availability of large-scale, historical road network data, which will fill an important gap in the geospatial data landscape. As such data is scarce, the quantitative evaluation of the extracted road networks is difficult. Therefore, we employed two types of reference data: historical built-up areas, and road presence-absence data obtained from manual map interpretation. The results suggest relatively high levels of agreement, increasing over time (e.g., Table 3). Observed disagreement (e.g., lower levels of accuracy for specific map sheets or points in time) likely represents a combined effect of misclassification and the incompatibility of the reference data used (i.e., presence of a road may not always indicate the presence of a built-up structure, or vice-versa). In addition to that, map quality affects the extraction results, as observed in the Mobile, Alabama, map which exhibits a very unique, rough cartographic style as a black-and-white print (Fig. 10).

As a result of misclassification, individual road segments of the extracted historical road network may be missing (omission errors) or may be spatially isolated (likely to represent commission errors). While such imperfect results can be used to identify and quantify urban growth and land development patterns, measured through the lens of the road network (e.g., Tables 4 and 5), topological or connectivity-based analyses conducted on the extracted historical road networks may be biased and require further refinement of the results. Such refinements could involve topological checks and heuristics to ensure the connectedness of the extracted road networks, integrated in the binarization process to "separate" historical from more recent road network segments (Section 2.2.4). This will enhance the usability of the extracted historical road networks for topology-based analyses. Moreover, such a topology-informed extraction will also overcome a current limitation, which is the data-driven, and rigorous derivation of the ROI threshold (using ck-means clustering) that may be the cause for some of the observed misclassifications.

In summary, the presented method represents an innovative example of spatial data integration and highlights how we can gain knowledge about past landscapes by leveraging color information harvested from increasingly digitally available historical maps at large spatial extents.

## 5. Conclusions

Herein, we propose a method that integrates contemporary, spatially explicit vector road network data and color information from historical topographic maps in order to reconstruct past road networks at high spatial detail, and in a fully automated manner. To our knowledge, this analysis represents the first large-scale study that "translates" color information from historical maps into quantitative indicators using road network segments as analytical units. This approach enables the direct measurement and quantification of land development and urban growth through a road network lens in a completely unsupervised manner. Thus, it constitutes an important step towards the fully automatic preservation of spatial-historical information on past landscapes contained in historical cartographic documents.

The proposed approach is robust to the choice of user-set parameters. It is unsupervised and thus, does not require any training data. It is computationally efficient and could be applied at country scales at a feasible level of data processing effort. Moreover, the results of the sensitivity analysis (Fig. 17) suggest that through modification of the parameters used to establish the ROI, the extraction can be made more efficient without affecting the results.

Despite the observed, high levels of agreement with a manually labelled reference database, as well as acceptable agreement with historical built-up areas, there are a few shortcomings of the proposed approach. (A) Our method is based on the contemporary road network as the analysis universe, and allows for measuring road network growth over time, but not the shrinkage of road networks (i.e., roads that have disappeared over time). While shrinkage of transportation infrastructure is common in the case of railroad networks in the US (Levinson 2005), road network shrinkage is not common and typically neglected in scientific studies (e.g., Meijer et al. 2018). (B) Our method is not capable to discriminate between roads and other linear features depicted in historical maps, such as railroads, contour lines, map graticule, or administrative borders. Future work could make use of additional color information, or of a supervised classification approach to tackle this problem. Future work will also include testing alternative clustering techniques (e.g., Schubert et al. 2017, Liu et al. 2008) and binarization strategies for the continuous ROI measure (e.g., Otsu 1979, Sabo et al. 2013), or a-posteriori refinement strategies, such as topological assessments to identify disconnected road segments. Moreover, the proposed method could be employed as a labeled data generator for the automated training data generation by sampling from both sides of the ROI distributions. These training data could then be input to computer-vision based road recognition models using convolutional neural networks or similar approaches (Saeedimoghaddam & Stepinski 2020, Can et al. 2021, Ekim et al. 2021). Such a two-staged approach would also overcome the shortcoming of ignoring road network shrinkage.

The availability of spatially explicit, historical road network data over large geographic (and temporal) extents will enhance a variety of research directions, such as road network scaling (e.g., Strano et al. 2017), urban growth simulation (e.g., Zhao et al. 2014, Zhao et al. 2016), economic studies related to the road network (e.g., Iacono & Levinson 2016, Miatto et al. 2017) and studies on the co-evolution of road networks and other components of the built environment (Achibet et al. 2014). As our approach does not extract historical road geometries from raster maps directly, but annotates contemporary vector data, the results, once topologically cleaned, could be directly input to topology- and graph based systems for change analysis (e.g., Lohfink et al. 2010, Shbita et al. 2020).

Concluding, the presented method and results illustrate how the integration of multi-source geospatial data allows for the generation of enriched, novel data infrastructure, constituting a fundamental prerequisite to enhance our knowledge of the long-term evolution of contemporary geographic phenomena. Ultimately, a thorough understanding of the long-term development of urban and rural transportation infrastructure will enable a more informed urban and regional planning, and make future transportation infrastructure more efficient, resilient, and sustainable.

**Funding:** This work was supported in part by the National Science Foundation through the University of Colorado Boulder under Grant IIS 1563933, and in part by the National Science Foundation through the University of Southern California under Grant IIS 1564164.

**Author Contributions:** J.U. and S.L. designed the study. J.U. acquired and processed the data. J.U. performed analysis and visualized the results. J.U., S.L., Y.Y.C., and C.K. wrote the paper.

**Declarations of interest:** None.

# 6. Appendix

**Appendix Table A1.** Binary classification accuracy assessment of the ck-means clusters against manually created reference data and reference labels based on the built-up area fraction per road buffer, including urban and rural strata for each scenario (Prec = Precision, Rec = Recall).

| Referene data source: | | | | Reference labels from built-up fraction >0% | | | | | | Manually created reference labels | | | | | |
|---|---|---|---|---|---|---|---|---|---|---|---|---|---|---|---|
| Cluster analysis setup: | | | | Per study area | | | Per map sheet | | | Per study area | | | Per map sheet | | |
| Weighting scheme | Study area | Year | Stratum | Prec | Rec | F1 | Prec | Rec | F1 | Prec | Rec | F1 | Prec | Rec | F1 |
| road length | AL | 1920 | overall | 0.12 | 0.43 | 0.19 | 0.17 | 0.43 | 0.24 | 0.67 | 0.50 | 0.58 | 0.69 | 0.36 | 0.48 |
| | AL | 1920 | rural | 0.07 | 0.54 | 0.12 | 0.11 | 0.54 | 0.19 | 0.70 | 0.59 | 0.64 | 0.77 | 0.39 | 0.52 |
| | AL | 1920 | urban | 0.30 | 0.36 | 0.33 | 0.28 | 0.36 | 0.31 | 0.58 | 0.30 | 0.40 | 0.53 | 0.30 | 0.38 |
| | CA | 1900 | overall | 0.42 | 0.73 | 0.53 | 0.40 | 0.57 | 0.47 | 0.52 | 0.68 | 0.59 | 0.58 | 0.62 | 0.60 |
| | CA | 1900 | rural | 0.36 | 0.88 | 0.51 | 0.29 | 0.47 | 0.36 | 0.41 | 0.81 | 0.54 | 0.48 | 0.63 | 0.54 |
| | CA | 1900 | urban | 0.51 | 0.61 | 0.56 | 0.52 | 0.64 | 0.57 | 0.68 | 0.59 | 0.63 | 0.69 | 0.62 | 0.65 |
| | CA | 1950 | overall | 0.74 | 0.58 | 0.65 | 0.79 | 0.57 | 0.66 | 0.89 | 0.67 | 0.77 | 0.83 | 0.57 | 0.67 |
| | CA | 1950 | rural | 0.65 | 0.63 | 0.64 | 0.79 | 0.64 | 0.70 | 0.91 | 0.82 | 0.87 | 0.77 | 0.58 | 0.66 |
| | CA | 1950 | urban | 0.78 | 0.57 | 0.66 | 0.79 | 0.54 | 0.64 | 0.88 | 0.60 | 0.72 | 0.86 | 0.56 | 0.68 |
| | NY | 1900 | overall | 0.63 | 0.64 | 0.63 | 0.67 | 0.64 | 0.65 | 0.75 | 0.77 | 0.76 | 0.80 | 0.77 | 0.79 |
| | NY | 1900 | rural | 0.63 | 0.72 | 0.67 | 0.66 | 0.72 | 0.69 | 0.81 | 0.84 | 0.83 | 0.85 | 0.84 | 0.84 |
| | NY | 1900 | urban | 0.62 | 0.49 | 0.55 | 0.70 | 0.49 | 0.58 | 0.60 | 0.60 | 0.60 | 0.68 | 0.61 | 0.64 |
| | NY | 1930 | overall | 0.83 | 0.72 | 0.77 | 0.84 | 0.72 | 0.77 | 0.84 | 0.93 | 0.89 | 0.84 | 0.93 | 0.88 |
| | NY | 1930 | rural | 0.80 | 0.73 | 0.76 | 0.80 | 0.73 | 0.76 | 0.84 | 1.00 | 0.91 | 0.84 | 1.00 | 0.91 |
| | NY | 1930 | urban | 0.90 | 0.69 | 0.78 | 0.93 | 0.70 | 0.80 | 0.85 | 0.82 | 0.83 | 0.84 | 0.79 | 0.82 |
| | NY | 1950 | overall | 0.77 | 0.78 | 0.78 | 0.76 | 0.73 | 0.74 | 0.97 | 0.93 | 0.95 | 0.97 | 0.88 | 0.92 |
| | NY | 1950 | rural | 0.70 | 0.93 | 0.80 | 0.67 | 0.83 | 0.74 | 0.97 | 0.97 | 0.97 | 0.97 | 0.89 | 0.93 |
| | NY | 1950 | urban | 0.92 | 0.63 | 0.75 | 0.92 | 0.63 | 0.75 | 0.97 | 0.85 | 0.90 | 0.96 | 0.85 | 0.90 |
| instance-based | AL | 1920 | overall | 0.19 | 0.36 | 0.24 | 0.22 | 0.36 | 0.27 | 0.67 | 0.39 | 0.49 | 0.61 | 0.30 | 0.41 |
| | AL | 1920 | rural | 0.09 | 0.50 | 0.15 | 0.14 | 0.50 | 0.22 | 0.73 | 0.53 | 0.62 | 0.71 | 0.33 | 0.45 |
| | AL | 1920 | urban | 0.25 | 0.33 | 0.29 | 0.25 | 0.33 | 0.29 | 0.63 | 0.32 | 0.43 | 0.56 | 0.29 | 0.38 |
| | CA | 1900 | overall | 0.48 | 0.70 | 0.57 | 0.50 | 0.70 | 0.58 | 0.63 | 0.58 | 0.60 | 0.65 | 0.58 | 0.62 |
| | CA | 1900 | rural | 0.44 | 0.88 | 0.58 | 0.45 | 0.63 | 0.53 | 0.38 | 0.86 | 0.52 | 0.45 | 0.71 | 0.56 |
| | CA | 1900 | urban | 0.49 | 0.67 | 0.57 | 0.51 | 0.71 | 0.59 | 0.69 | 0.56 | 0.62 | 0.68 | 0.57 | 0.62 |
| | CA | 1950 | overall | 0.76 | 0.61 | 0.68 | 0.77 | 0.61 | 0.68 | 0.86 | 0.62 | 0.72 | 0.84 | 0.59 | 0.69 |
| | CA | 1950 | rural | 0.70 | 0.58 | 0.64 | 0.80 | 0.67 | 0.73 | 0.90 | 0.75 | 0.82 | 0.80 | 0.67 | 0.73 |
| | CA | 1950 | urban | 0.77 | 0.61 | 0.68 | 0.77 | 0.60 | 0.67 | 0.85 | 0.60 | 0.71 | 0.84 | 0.59 | 0.69 |
| | NY | 1900 | overall | 0.70 | 0.60 | 0.65 | 0.73 | 0.61 | 0.67 | 0.78 | 0.72 | 0.75 | 0.81 | 0.74 | 0.77 |
| | NY | 1900 | rural | 0.60 | 0.82 | 0.69 | 0.64 | 0.82 | 0.72 | 0.73 | 0.92 | 0.81 | 0.79 | 0.92 | 0.85 |
| | NY | 1900 | urban | 0.73 | 0.56 | 0.63 | 0.76 | 0.58 | 0.65 | 0.80 | 0.68 | 0.73 | 0.82 | 0.70 | 0.76 |
| | NY | 1930 | overall | 0.85 | 0.71 | 0.77 | 0.86 | 0.72 | 0.78 | 0.89 | 0.83 | 0.86 | 0.89 | 0.83 | 0.86 |
| | NY | 1930 | rural | 0.76 | 0.81 | 0.79 | 0.76 | 0.81 | 0.79 | 0.88 | 1.00 | 0.94 | 0.88 | 1.00 | 0.94 |
| | NY | 1930 | urban | 0.87 | 0.69 | 0.77 | 0.89 | 0.70 | 0.78 | 0.89 | 0.79 | 0.84 | 0.89 | 0.79 | 0.84 |
| | NY | 1950 | overall | 0.86 | 0.66 | 0.75 | 0.84 | 0.65 | 0.73 | 0.97 | 0.84 | 0.90 | 0.96 | 0.83 | 0.89 |
| | NY | 1950 | rural | 0.71 | 0.92 | 0.80 | 0.69 | 0.85 | 0.76 | 0.94 | 0.94 | 0.94 | 0.94 | 0.88 | 0.91 |
| | NY | 1950 | urban | 0.90 | 0.62 | 0.73 | 0.89 | 0.62 | 0.73 | 0.98 | 0.82 | 0.89 | 0.97 | 0.82 | 0.89 |

**Appendix Table A2. Multi-temporal road network statistics based on cross-tabulation of cluster labels across time, within strata of urban and rural roads, and for both clustering strategies.**

| Clustering strategy | Study area | T1 | T2 | Stratum | km road | | | | [%] of contemporary road length | | | |
|---|---|---|---|---|---|---|---|---|---|---|---|---|
| | | | | | not existent | persistent | newly built | disappeared | not existent | persistent | newly built | disappeared |
| Clustering per study area | NY | 1900 | 1930 | overall | 3,625 | 3,756 | 2,522 | 605 | 34.50 | 35.75 | 24.00 | 5.75 |
| | NY | 1900 | 1930 | urban | 2,490 | 1,747 | 1,391 | 353 | 41.64 | 29.21 | 23.26 | 5.90 |
| | NY | 1900 | 1930 | rural | 1,135 | 2,009 | 1,131 | 252 | 25.07 | 44.39 | 24.98 | 5.56 |
| | NY | 1930 | 1950 | overall | 3,399 | 5,664 | 825 | 613 | 32.36 | 53.94 | 7.86 | 5.84 |
| | NY | 1930 | 1950 | urban | 2,312 | 2,824 | 526 | 311 | 38.70 | 47.29 | 8.81 | 5.20 |
| | NY | 1930 | 1950 | rural | 1,087 | 2,840 | 299 | 303 | 24.01 | 62.71 | 6.60 | 6.69 |
| | NY | 1900 | 1950 | overall | 3,379 | 3,727 | 2,762 | 633 | 32.18 | 35.49 | 26.31 | 6.03 |
| | NY | 1900 | 1950 | urban | 2,272 | 1,746 | 1,605 | 350 | 38.05 | 29.23 | 26.87 | 5.85 |
| | NY | 1900 | 1950 | rural | 1,107 | 1,981 | 1,157 | 283 | 24.44 | 43.75 | 25.56 | 6.25 |
| | CA | 1900 | 1950 | overall | 12,966 | 10,544 | 9,253 | 4.578 | 34.72 | 28.24 | 24.78 | 12.26 |
| | CA | 1900 | 1950 | urban | 8,397 | 5,495 | 6,080 | 2.588 | 37.22 | 24.36 | 26.95 | 11.47 |
| | CA | 1900 | 1950 | rural | 4,569 | 5,049 | 3,173 | 1.990 | 30.91 | 34.16 | 21.47 | 13.46 |
| Clustering per map sheet | NY | 1900 | 1930 | overall | 3,317 | 3,772 | 2,277 | 503 | 33.61 | 38.22 | 23.07 | 5.10 |
| | NY | 1900 | 1930 | urban | 2,417 | 1,720 | 1,307 | 346 | 41.75 | 29.70 | 22.57 | 5.97 |
| | NY | 1900 | 1930 | rural | 900 | 2,053 | 970 | 158 | 22.06 | 50.30 | 23.77 | 3.87 |
| | NY | 1930 | 1950 | overall | 3,105 | 5,572 | 716 | 477 | 31.46 | 56.45 | 7.25 | 4.84 |
| | NY | 1930 | 1950 | urban | 2,234 | 2,767 | 528 | 260 | 38.60 | 47.79 | 9.12 | 4.48 |
| | NY | 1930 | 1950 | rural | 871 | 2,805 | 187 | 218 | 21.34 | 68.73 | 4.59 | 5.33 |
| | NY | 1900 | 1950 | overall | 3,082 | 3,775 | 2,513 | 501 | 31.22 | 38.25 | 25.46 | 5.08 |
| | NY | 1900 | 1950 | urban | 2,175 | 1,746 | 1,549 | 319 | 37.58 | 30.17 | 26.75 | 5.50 |
| | NY | 1900 | 1950 | rural | 906 | 2,028 | 964 | 182 | 22.21 | 49.70 | 23.62 | 4.47 |
| | CA | 1900 | 1950 | overall | 13,026 | 10,142 | 8,097 | 4.096 | 36.84 | 28.68 | 22.90 | 11.58 |
| | CA | 1900 | 1950 | urban | 8,575 | 5,880 | 5,544 | 2.367 | 38.34 | 26.29 | 24.79 | 10.59 |
| | CA | 1900 | 1950 | rural | 4,451 | 4,262 | 2,553 | 1.728 | 34.25 | 32.80 | 19.65 | 13.30 |

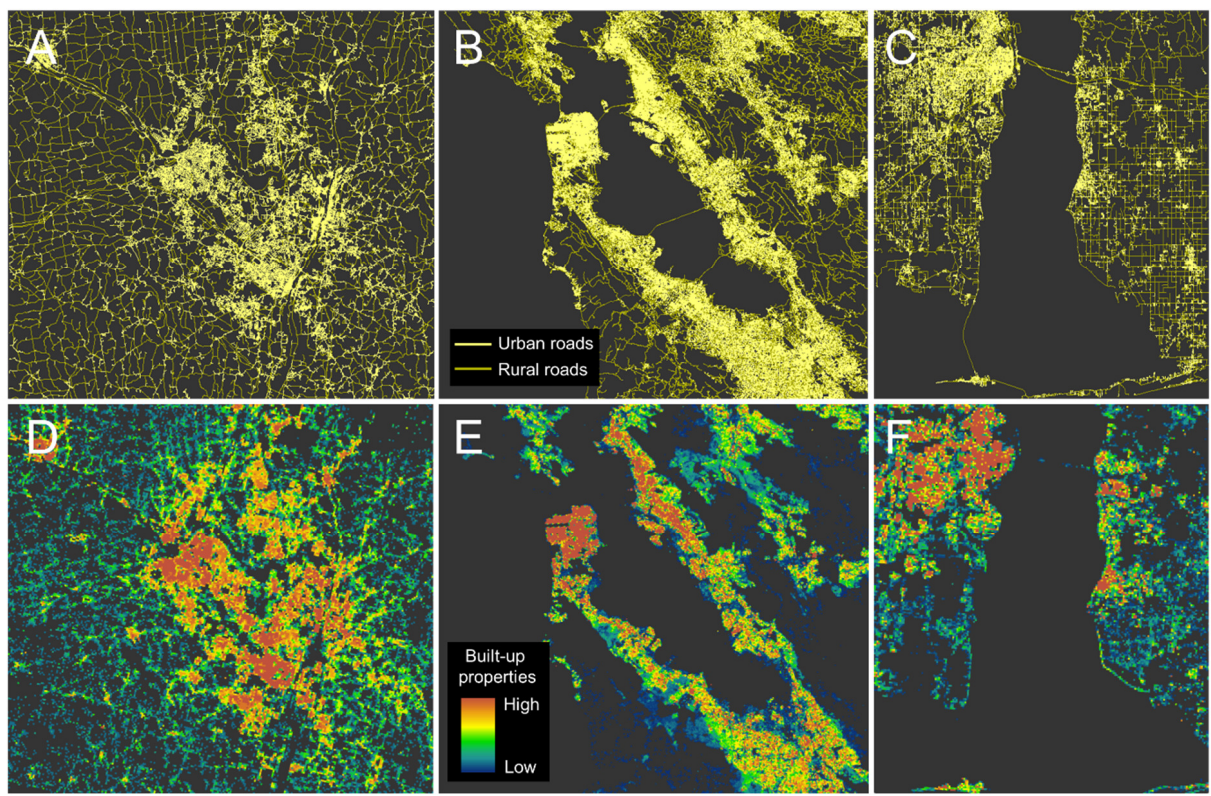

**Appendix Figure A1. Top row: Visualization of segment-length based strata of roads, roughly classifying roads into "urban" roads (segment length < 90th percentile) and "rural" roads (segment length >= 90th percentile). (a) Albany (NY), (b) Bay area (CA), and (c) Mobile bay (AL). Percentiles are based on road**

segment length distributions per study area. While we did not derive this threshold empirically, we visually cross-compared the extents of the urban road segments with a built-up property (BUPR) density surface from the Historical Settlement Data Compilation for the US (HISDAC-US, Uhl et al. 2021b), shown in panels (d)-(f). The BUPR surface maps built-up properties in 2016, derived from cadastral data, and is a rough proxy measure for building density. We find high levels of agreement between the "urban" roads and high-density settlements.

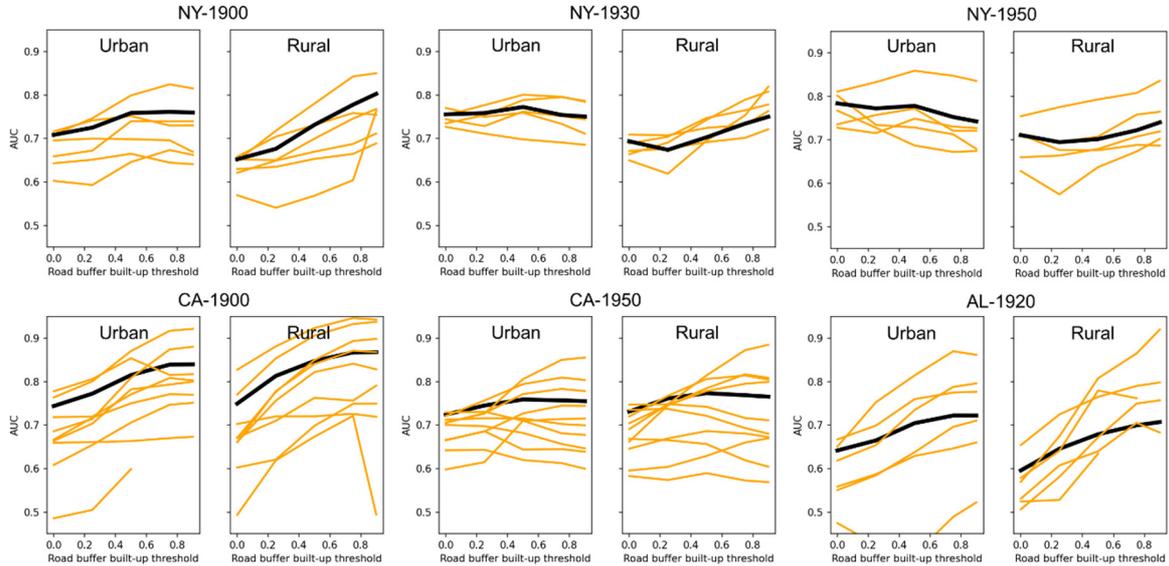

**Appendix Figure A2.** ROC analysis results of the continuous road overlap indicator against discrete, for different thresholds applied to the building-based reference data. Different thresholds were applied to the built-up area fraction within road buffer areas, for generating optimistic and conservative reference labels, i.e., we assume that a road exists if HISDAC-US reports at least one building in the neighborhood of the street, (threshold>0), or only if more than 90% of the area in the neighborhood of the street are developed (threshold>0.9).

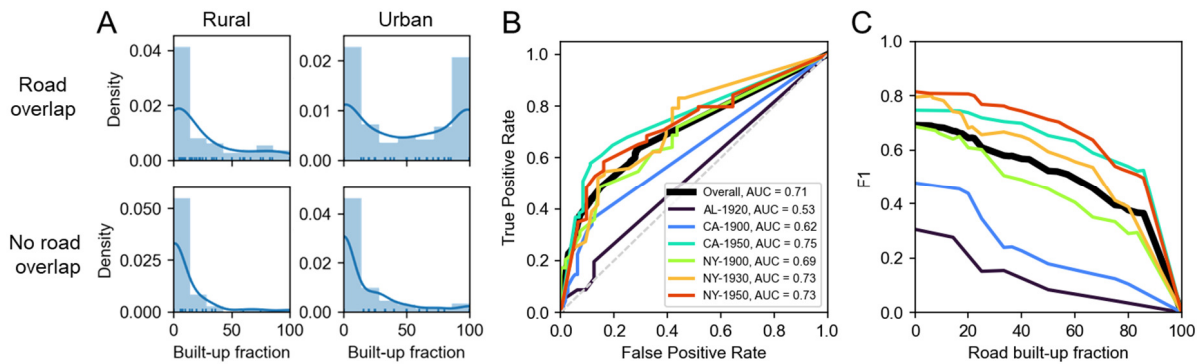

**Appendix Figure A3.** Cross-comparison of manually and automatically generated reference data. Agreement between binary labels (road exists / does not exist in map) vs. continuous built-up area fractions in the proximity of the roads.

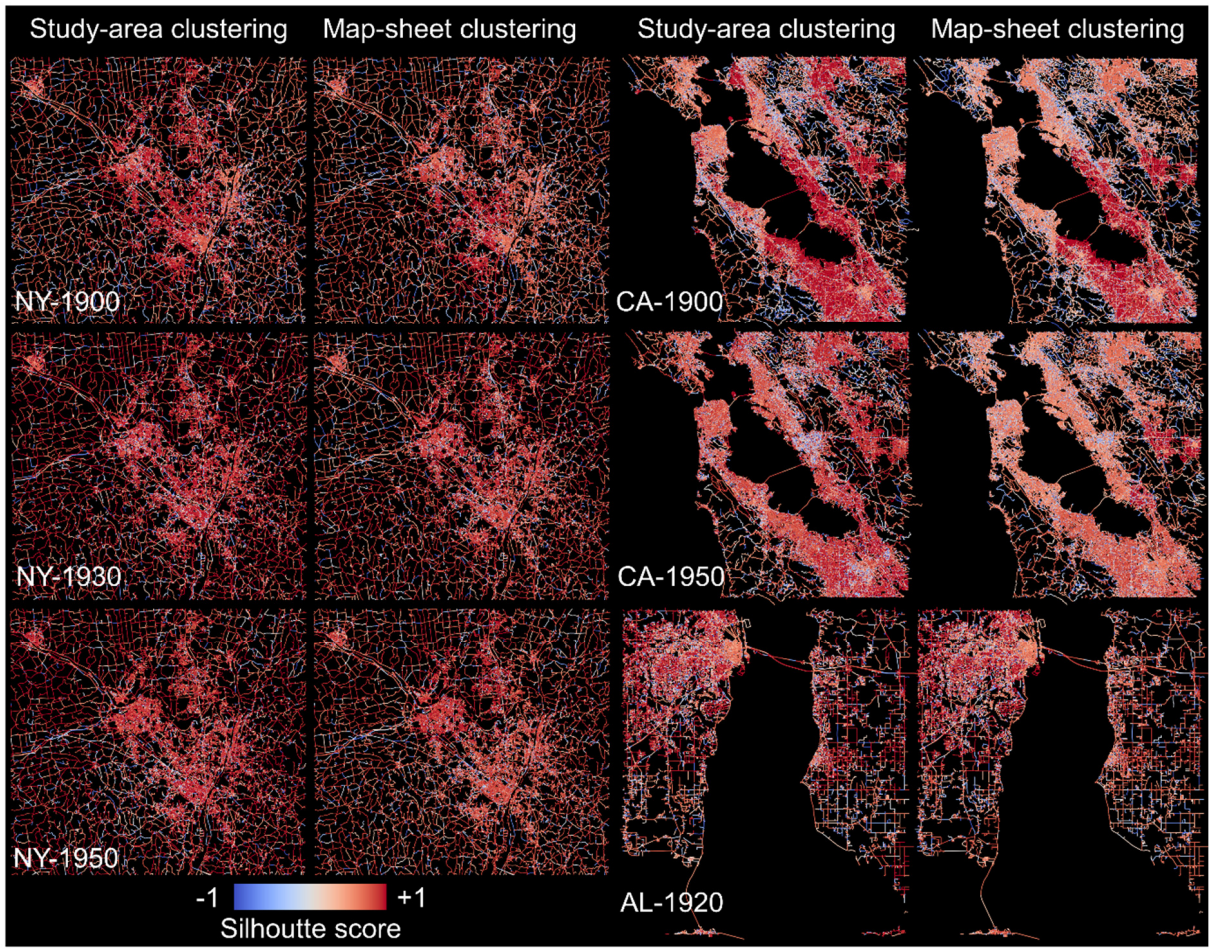

**Appendix Figure A4. Silhouette scores for each road segment from ck-means clustering for each study area, point in time, and clustering strategy.**

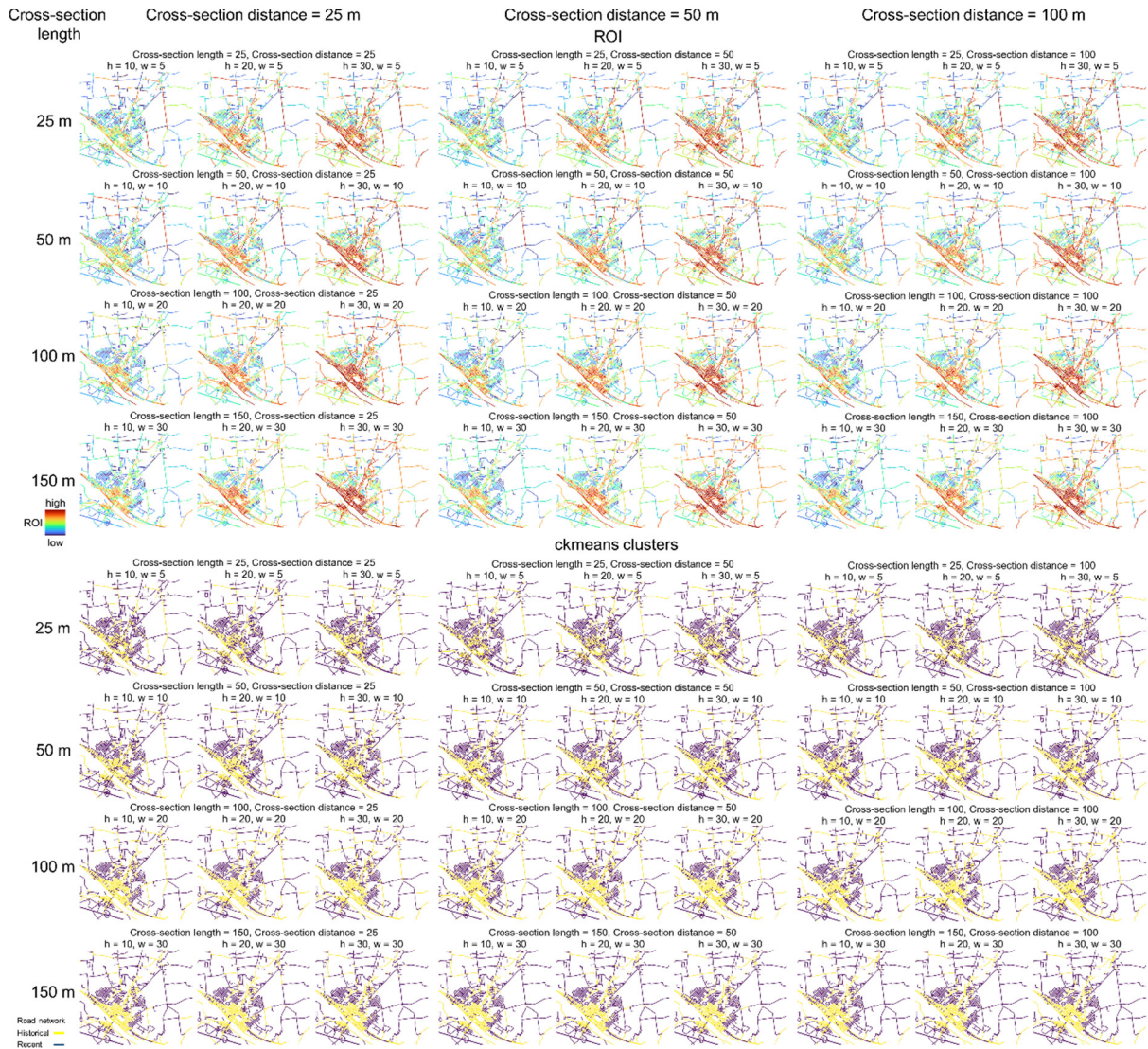

**Appendix Figure A5.** Sensitivity analysis results varying the cross-section length, cross section distance used to sample color information from the historical maps, and the width *w* and height *h* of the axial images used to construct the road overlap metric. Shown are the ROI (upper part) and corresponding ck-means clusters (lower part), based on a 1895 map of Amsterdam (NY).